%% file: ms.tex
\documentclass[preprint,journal]{vgtc}       

\ifpdf
  \pdfoutput=1\relax                   
  \usepackage{graphicx}                
  \DeclareGraphicsExtensions{.pdf,.png,.jpg,.jpeg} 
\else
  \usepackage{graphicx}                
  \DeclareGraphicsExtensions{.eps}     
\fi%

\graphicspath{{figures/}{pictures/}{images/}{./}} 

\usepackage{microtype}                 
\PassOptionsToPackage{warn}{textcomp}  
\usepackage{textcomp}                  
\usepackage{mathptmx}                  
\usepackage{times}                     
\usepackage{cite}                      
\usepackage{tabu}                      
\usepackage{booktabs}                  

\usepackage{algorithm}
\usepackage{algpseudocode}

\newcommand{\etal}{\textit{et al}.}

\onlineid{1380}

\vgtccategory{Research}
\vgtcpapertype{algorithm/technique}

\title{Visualizing the Passage of Time with Video Temporal Pyramids}

\author{Melissa E. Swift, \textit{Member, IEEE}, Wyatt Ayers, Sophie Pallanck, and Scott Wehrwein}
\authorfooter{
\item
 Melissa E. Swift conducted this research while at Western Washington University and is currently with Pacific Northwest National Laboratory. E-mail: melissa.swift@pnnl.gov.
\item
 Wyatt Ayers is at Western Washington University. E-mail: ayersw2@wwu.edu.
\item
 Sophie Pallanck was at Western Washington University. E-mail: sophierosepallanck@gmail.com.
\item
 Scott Wehrwein is at Western Washington University. E-mail: scott.wehrwein@wwu.edu.
}

\shortauthortitle{Swift \MakeLowercase{\textit{et al.}}: Visualizing the Passage of Time with Video Temporal Pyramids}

\input{sections/0_abstract}

\keywords{Time, time-frequency, video visualization, multi-scale, webcam}

\teaser{
  \centering
  \includegraphics[width=\linewidth]{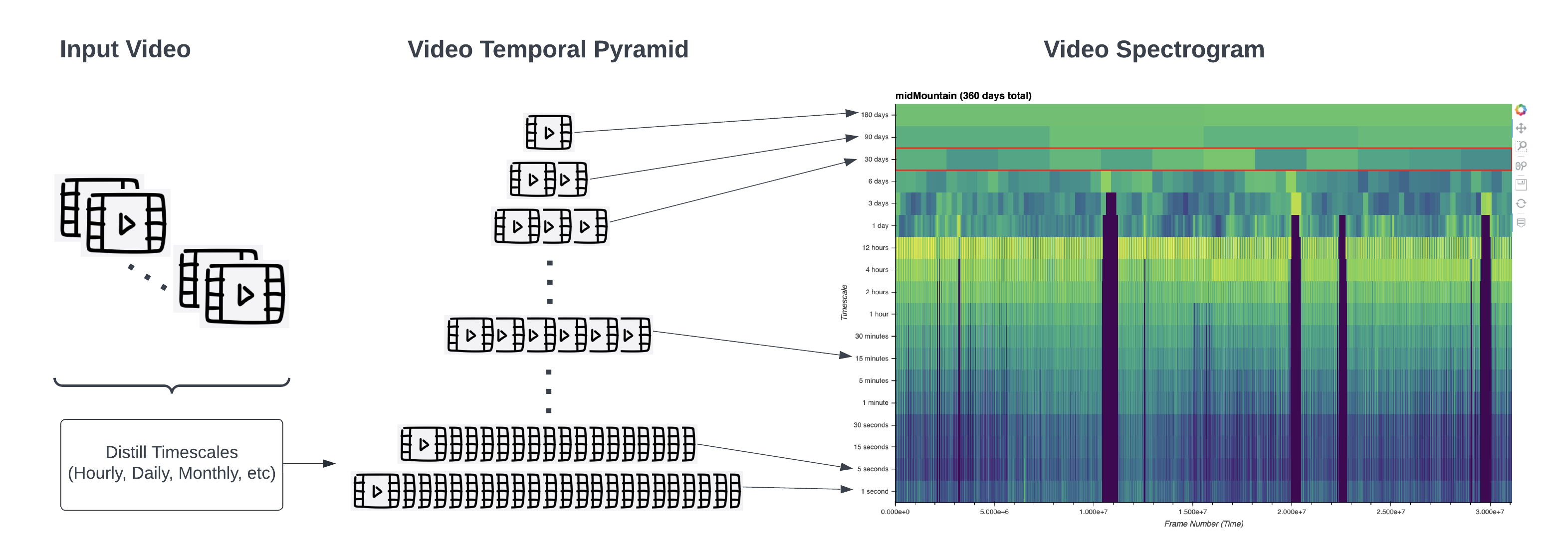}
  \caption{Given months or years of recorded webcam footage, our approach builds a Video Temporal Pyramid consisting of different length shorter videos, each of which visualizes the events happening at a particular timescale. Our Video Spectrogram is a visualization for the pyramid that provides both overview and drill down functionality to aid in interactive exploration. 
}
  \label{fig:teaser}
}

\vgtcinsertpkg

\begin{document}


\firstsection{Introduction}

\maketitle
\input{sections/1_intro}

\input{sections/2_related}
\input{sections/3_pyramid}

\input{sections/4_spectrogram}

\input{sections/5_results}

\input{sections/6_discussion}

\acknowledgments{
This work was supported in part by NASA Award NNX15AJ98H under the Washington NASA Space Grant Consortium, and in part by the National Science Foundation under Grant No. 2105372.  The Washington NASA Space Grant Consortium is funded by the NASA Office of Stem Engagement.  Any opinions, findings, and conclusions or recommendations expressed in this material are those of the author(s) and do not necessarily reflect the views of NASA or the NSF.

The authors wish to thank Ann Tseng and Richie Mohan for their early contributions.}

\bibliographystyle{abbrv-doi-hyperref-narrow}

\bibliography{ms}
\end{document}

%% file: sections/0_abstract.tex
\abstract{
What can we learn about a scene by watching it for months or years? A video recorded over a long timespan will depict interesting phenomena at multiple timescales, but identifying and viewing them presents a challenge. The video is too long to watch in full, and some things are too slow to experience in real-time, such as glacial retreat or the gradual shift from summer to fall. Timelapse videography is a common approach to summarizing long videos and visualizing slow timescales. However, a timelapse is limited to a single chosen temporal frequency, and often appears flickery due to aliasing. Also, the length of the timelapse video is directly tied to its temporal resolution, which necessitates tradeoffs between those two facets. In this paper, we propose Video Temporal Pyramids, a technique that addresses these limitations and expands the possibilities for visualizing the passage of time. Inspired by spatial image pyramids from computer vision, we developed an algorithm that builds video pyramids in the temporal domain. Each level of a Video Temporal Pyramid visualizes a different timescale; for instance, videos from the monthly timescale are usually good for visualizing seasonal changes, while videos from the one-minute timescale are best for visualizing sunrise or the movement of clouds across the sky. To help explore the different pyramid levels, we also propose a Video Spectrogram to visualize the amount of activity across the entire pyramid, providing a holistic overview of the scene dynamics and the ability to explore and discover phenomena across time and timescales. To demonstrate our approach, we have built Video Temporal Pyramids from ten outdoor scenes, each containing months or years of data. We compare Video Temporal Pyramid layers to naive timelapse and find that our pyramids enable alias-free viewing of longer-term changes. We also demonstrate that the Video Spectrogram facilitates exploration and discovery of phenomena across pyramid levels, by enabling both overview and detail-focused perspectives. \href{https://fw.cs.wwu.edu/~wehrwes/TemporalPyramids}{\texttt{https://fw.cs.wwu.edu/\(\sim \)wehrwes/TemporalPyramids}}
} 

%% file: sections/1_intro.tex
The world around us is constantly changing at many speeds at once, but the human visual system can only perceive a narrow range of dynamic phenomena in real time. Some things move too slowly for us to register, such as a glacier flowing, and some things move too quickly for us to register, such as a bee's wings in flight. Though we cannot see these motions as they occur, we can visualize them after the fact. For fast events, we can use a high-speed camera and slow down the footage to a more human-friendly speed (i.e., slow-motion). 
For slow events, the most common method of visualizing these changes is timelapse photography, where frames are taken at a regular intervals over time and then assembled into a video that plays much faster. 
The key observation that motivates our work is that many scenes exhibit interesting phenomena at multiple timescales: in a single scene, we might be able to observe foot and vehicle traffic on a road, movement of clouds in the sky, diurnal changes in illumination, and a building being constructed over the course of months or years.

A natural way to capture these multi-timescale phenomena is to begin with an input video with sufficiently high framerate to capture the fastest-moving phenomena. However, months of raw video cannot be viewed in a reasonable amount of time, so we might subsample it to create a series of timelapse videos that show different rates (e.g., one frame per minute, one frame per hour, etc.). However, straightforward timelapse sampling exhibits aliasing due to high-frequency content. Consider sampling one frame per month; although longer-term changes happening at or around a months-long timescale will be naturally viewable, shorter-term changes, such as a person that happened to be walking through the scene at the moment a frame was sampled, will appear as a distracting single-frame blip. This paper proposes (1) Video Temporal Pyramids, a more principled, alias-free approach for visualizing changes at different timescales; and (2) the Video Spectrogram, a visualization tool for navigating and exploring the pyramids.

The algorithm that forms the basis for creating a Video Temporal Pyramid takes inspiration from the common image processing techniques of Gaussian and Laplacian image pyramids, but applied to the temporal domain. The result is a collection of new videos, each of which distills the changes happening at a particular timescale  (e.g., hourly, daily, monthly, yearly). Though each pyramid level is similar to a timelapse with a particular sampling rate, they feature a smoother viewing experience with no aliasing or flickering effects. 

The Video Temporal Pyramid captures information about the changes over many timescales, but the volume of video data is (approximately $2\times$) larger than the original. To help a user navigate and explore the pyramid and surface more information about events and patterns in the scene, we propose a visualization tool called the Video Spectrogram. We quantify the magnitude of changes happening at each time and in each timescale and plot those data as a heatmap, analogous to the spectrogram used in audio processing \cite{cohen1995time}. The resulting spectrogram plots time vs. timescale, showing clear patterns for strong cyclic changes such as day/night and seasonal periodicity. Anomalies such as significant weather events and corrupted data can also be discovered. The Video Spectrogram facilitates connecting the video footage to specific dates and times when events occurred. This enables the user to quickly do an overview scan and then drill down to lower timescales to view more details for a particular day or time, in keeping with Shneiderman's information seeking mantra \cite{schneiderman1996mantra}. This is key to making the large volume of video data manageable without arbitrarily discarding information. 

To validate our contributions we have processed multiple long-duration webcam datasets of diverse outdoor scenes, including a construction site, a ski slope, and a mountain lake, among others. The time periods covered by our datasets range from 1 month to 16 years, with base temporal resolutions ranging from 30 frames per second to 1 frame per hour (\autoref{fig:dataset_list}). In our exploration of these datasets, especially in direct comparison to a timelapse baseline, we found that our pyramid videos and spectrogram tool allowed us to rapidly learn a lot of detailed information about each scene, from how the seasons changed all the way down to the exact time a particular object appeared in the scene. Please refer to our supplementary materials to view selected Video Temporal Pyramid videos from each of our datasets as well as a demo of the Video Spectrogram. 

%% file: sections/2_related.tex
\section{Background and Related Work}
The proposed Video Temporal Pyramid and Video Spectrogram are closely related to work in several subdisciplines. This section provides a brief overview of the most relevant.

\subsection{Timelapse and Related Techniques}
Timelapse has been in existence since the late 1800's \cite{wikipedia2022timelapse} and is a popular way to visualize the passage of time on a small scale (\textit{e.g.}, a pineapple rotting \cite{temponaut2021vid}) or a large scale (e.g., Google Earth Timelapse \cite{google2022timelapse}). Of particular interest, Martin-Brualla \etal \cite{martinbrualla2015timelapse} drew from internet imagery to create years-long timelapse videos, but the noise from internet photos captured by different cameras and at different times requires heavy smoothing so that shorter-term changes are not visible and long-term changes can be hard to detect. 

Several techniques have been proposed to smooth out the aliasing artifacts that result from timelapse sampling after the fact. 
With consumer video applications (\textit{e.g.}, hyperlapse and timelapse) in mind, Zhang \etal \cite{zhang2017photometric} propose a method for smoothing transitions between frames in temporally subsampled videos. Whereas their method interpolates and smooths after subsampling, our method necessitates fewer modeling assumptions because we smooth discontinuities before temporal subsampling; starting with the densely sampled input also allows us to produce smooth visualizations of any timescale. Their method is also tested only on videos that span minutes or hours of time, and are subsampled to seconds or minutes, whereas ours is designed to work with years-long video streams. Finally, their method is also more computationally expensive, operating around 0.5 frames per second, whereas ours runs at 6 frames per second. While both methods can benefit from parallelization, this is nonetheless a significant difference when considering datasets such as ours that have on the order of 1 billion frames. Details on runtime and datasets can be found in the supplemental material.

In remote sensing, it is often desirable to visualize long-term changes related to a variety of phenomena caused by humans or natural events. The data is often of low temporal frequency, sometimes only before-and-after satellite pictures or landscape photographs taken far apart in time, such as the U.S. Geological Survey repeat photography project \cite{usgs2016repeat}. Animation techniques have been proposed which can help create a smooth transition between images. Lobo \etal \cite{lobo2019satellite} does this by simulating plausible intermediary frames, while Harrower \cite{harrower2001animation} provides the user with control over the spatial and temporal resolution to allow for optimal visualization of a given phenomenon. While our work shares the same goals of visualizing changes happening over long timescales, we work with datasets with high temporal resolution and do not rely on interpolation to smooth transitions. That said, interpolation methods such as Lobo \etal \cite{lobo2019satellite} could be complementary to ours as a possible way to fill in segments of missing data from our datasets.

\subsection{Temporal Resampling and Video Visualization}
Most existing techniques for video visualization are designed for videos no longer than a few hours, and their end goals often differ significantly from ours. In fact, over ten years ago, Borgo \etal \cite{borgo2011vidviz} published a survey of different video visualization techniques; while our work is related to many of the techniques they describe, the authors clearly assumed the use of relatively short videos. This section highlights some of the most closely related work in this area.

Various works adapt the frame rate, or temporal sampling rate, of a video over time based on its content \cite{zhou2014time, joshi2015hyperlapse}. The most closely related technique is Computational Timelapse \cite{bennett2007timelapse}, which uses temporal differences in video frames to dynamically speed up the frame rate when little is changing and slow it down when more changes are occurring. While this is effective as an automatic fast-forward tool, it requires a chosen output video length; furthermore, long-term changes appear choppy and broken up due to sudden changes in the frame rate.
Several works have proposed various non-axis-aligned manipulations of space-time cubes such as videos; Rav-Acha \etal \cite{rav-acha05evolving} explore this idea from in a graphics/vision context, while Bach \etal give a thorough visualization-oriented review of the possibilities in this space. These techniques are generally incompatible with cuboids with significantly longer time extent, such as months-long videos.

Video summarization techniques take another approach---rather than maintaining chronological and spatial continuity, these techniques attempt to find frames or clips that encapsulate the activity in a video. These techniques generally approach the problem by automatically detecting ``noteworthy'' frames or clips either by using unsupervised saliency-based approaches \cite{pritch2008nonchronological} or by using example-based learning \cite{zhang2016video, rochan2018video}. These techniques tend to focus on the real-time timescale and treat longer-term changes as noise; they also make automated decisions that may remove content of interest even at the real-time timescale. 

Video fast-forward techniques are closely related to the ideas supporting timelapse videos and often use frame-skipping. The disadvantages of timelapse were explored by Hoferlin \etal \cite{hoferlin2012ffvis}. Their evaluation of fast-forward techniques also included an interesting experiment with the use of temporal blending---similar in spirit to our temporal filtering method. However, their filtering approach does not generalize to more than one timescale. 

A few prior works have considered multiple timescales of activity in videos. Motion Denoising \cite{rubinstein2011motion} separates a video into short-term and long-term components. This technique produces excellent results, but handles only two timescales and is very computationally expensive, making it impractical for months-long videos, much less for years-long videos. Wehrwein \etal \cite{wehrwein2021scene} propose a method to composite clips from different timescales into a single ``scene summary,'' showing, for example, people walking, clouds moving, and the sun crossing the sky all at once. This method begins with a Gaussian temporal pyramid much like the one constructed as a side-effect of our Laplacian pyramid construction, then composites salient clips from different pyramid layers together; though multiple timescales are visualized at once, the vast majority of the lower pyramid levels are discarded from the final output. In contrast, we construct visualizations that assists in exploration of the whole dataset without any assumptions about which timescales or scene elements are of interest to the viewer.

\subsection{Interactive Video Exploration and Retrieval}
In contrast to the automated methods above, a separate category of prior work facilitates interactive browsing, exploration, and retrieval in video. Though this is more closely related to our task, most of these techniques are oriented towards retrieval of specific content rather than discovery, or towards improving the ability to scrub or seek in a shorter video.

The Video Browser Showdown \cite{rossetto2020vbs} is a yearly contest of video browsing tools designed to locate either a specific event or clip in a video, or locate all instances of an event or action. Tools that are successful on this task (e.g., \cite{kratochvil2020somhunter}) tend to leverage the fact that the sequence of interest is known \textit{a priori}, making them less useful for discovering unknown anomalies; for similar reasons, these tools are also unlikely to generalize well for the purpose of identifying structure at longer timescales.

Several techniques have been proposed to show a visual overview of an entire video sequence, or improve scrubbing. Gutwin \etal  \cite{gutwin2019spreadloading} proposed a spread-loading scheme to load frames at varying intervals when loading a streaming video to improve the scrubbing experience. They showed that this improved users' ability to seek to a particular point in the video quickly, but even with instantaneous availability of all frames, a months-long video would be tedious to explore by scrubbing. Barnes \etal \cite{barnes2010tapestries} create a continuous visual overview of automatically selected keyframes to assist in scrubbing around in the video, while Jackson \etal \cite{jackson2013panopticon} arrange short clips of a video in an animated grid so a user can shift their focus to any point in a video or watch a single thumbnail as it cycles through the entire video. These techniques work well for shorter clips, but their utility is limited by available screen size for longer-duration videos. 

From a visualization perspective, Romero \etal \cite{romero2008vizavis} also uses computer vision to analyze and visualize video volumes. In particular, their interface proposed a closely related heat map visualization called the Activity Table that is similar in spirit to our Video Spectrogram. The Activity Table displays aggregate motion computed using thresholded frame differences, closely related to the frame differences performed in the construction of our Laplacian Pyramids. Whereas they plot aggregate motion in a single (real-time) timescale across different spatial locations, we are interested in activity at multiple timescales and use the vertical axis of the heatmap to index temporal frequency instead of spatial location.

\subsection{Time Series Visualization} 
Although our work relates to the rich literature on time series visualization (e.g., \cite{aigner2011timevis, ali2019timecluster}), video has unique properties that benefit from domain-specific techniques. One notable example from this literature is the work by Cakmak \etal \cite{cakmak2021multiscale} which does visualize time-varying data at multiple timescales; their interface for traversing temporal scales and viewing summaries at different levels is loosely analogous to our Video Spectrogram, but is geared towards the very different domain of time-varying graph data.

\subsection{Pyramids and Spectrograms}
The pyramid computation component of our approach is directly adapted from image pyramid techniques from the computer vision literature. Our adaptation will be described in detail below. In particular, we compute Gaussian \cite{burt1981pyramids} and Laplacian \cite{burt1983laplacian} pyramids along the time dimension, in contrast with their traditional application to the spatial dimensions of images. The idea of generalizing image pyramid techniques to videos is not new; a related generalization of the Gaussian pyramid to video was proposed by \cite{finkelstein1996multiresolution} to send videos at variable resolutions in space and time over limited bandwidth network connections. Their approach resembles a Gaussian pyramid (in contrast to our Laplacian pyramid), operates across spatial and temporal dimensions, and is designed for efficient coding and variable-resolution transmission of videos under limited bandwidth. Our method aims to visualize longer-term changes with high fidelity. For the problem of human action recognition, numerous works propose variations of pyramids applied temporally for short (e.g., minutes-long) video clips \cite{shao2014actionrecog, lan2014tsp, wang2017action, wang2017actionpooling, zheng2019action}. To our knowledge, however, our method is the first to extend the temporal Laplacian pyramid concept to extremely long-duration videos with the intent to visualize long timescales. Finally, our Video Spectrogram tool is directly inspired by the idea of time-frequency representations like the spectrogram, which are commonly used in audio visualization and processing \cite{cohen1995time}.

%% file: sections/3_pyramid.tex
\section{Video Temporal Pyramids}\label{sec:pyramids}

Our approach is inspired by image pyramids from the computer vision literature, which allow for separation and manipulation of spatial frequencies in images. We generalize these same techniques to the temporal domain in videos in order to separate and manipulate temporal frequencies. Specifically, the core of our Video Temporal Pyramid approach is a Laplacian pyramid computed in the time dimension, approximating the output of a bank of band-pass filters applied pixel-wise across time. We first formally define the pyramid's construction, then discuss several adaptations for the domain of long, fixed-camera videos.

\subsection{Definition and Construction}
\begin{figure}
    \centering
    \includegraphics[width=\linewidth]{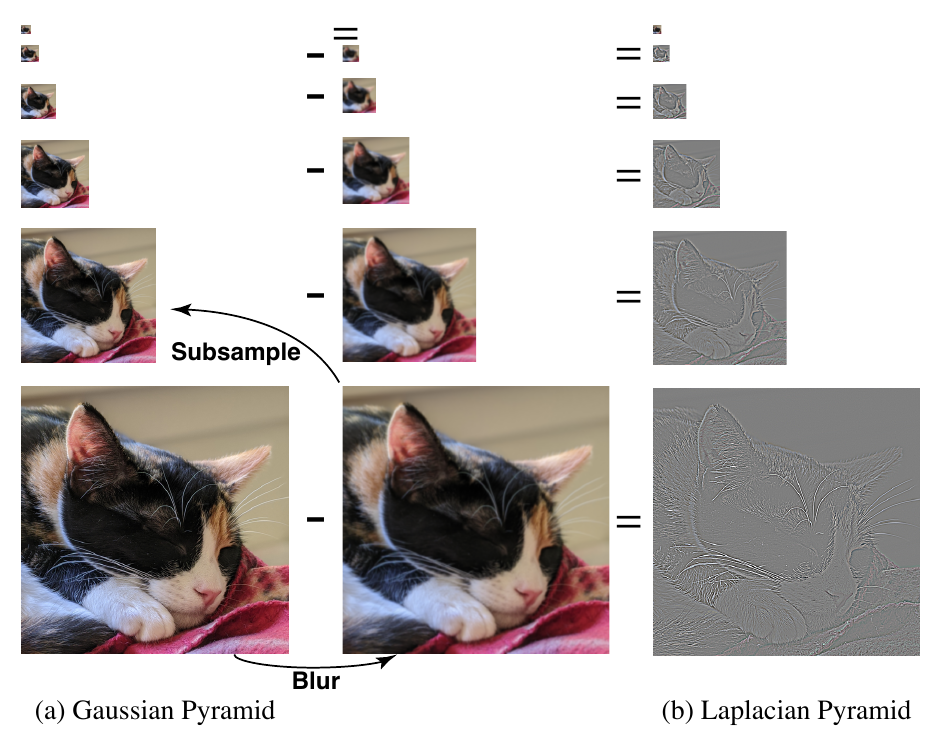}
    \caption{Traditionally, Gaussian and Laplacian pyramids are applied to both spatial dimensions of an image. The left column (a) shows a Gaussian pyramid each layer of which is blurred and subsampled from the prior one. Each level of the Laplacian pyramid (b) is a high-pass filtered version of the corresponding Gaussian level, computed by subtracting the blurred level from the original.}
    \label{fig:imgpyr}
\end{figure}

Image pyramids are a classical technique from the computer vision literature \cite{burt1981pyramids,burt1983laplacian}, widely used to apply image processing and computer vision algorithms at multiple scales or in a scale-invariant fashion. The most basic image pyramid is a Gaussian Pyramid \cite{burt1981pyramids} (\autoref{fig:imgpyr} (a)), constructed by repeatedly blurring then subsampling an image. Each subsequent level of the pyramid represents what remains after a low-pass filter is applied to the prior level. Each level of a Laplacian pyramid \cite{burt1983laplacian} (\autoref{fig:imgpyr} (b)) represents the result of a \textit{high-pass} filter applied to the prior level of the Gaussian pyramid, or equivalently a \textit{band-pass} filter applied to the original image. The resulting Laplacian pyramid levels contain narrow slices of spatial frequency content of the image, thereby resembling the output of a bank of band-pass filters.

Where Laplacian pyramids are traditionally used to isolate and manipulate \textit{spatial} frequency content of images, we instead apply the same procedure to the \textit{temporal} dimension of a video, leaving the spatial dimensions alone. A temporal analog to the Gaussian pyramid consists of videos that have been filtered and subsampled along the time dimension only. The Laplacian pyramid is also constructed analogously, by subtracting the temporally blurred video from the original. 
In principle, frames of the Laplacian temporal pyramid can be computed by subtracting the computed blur frames from the current level of the Gaussian pyramid before subsampling. In practice, we use the standard pyramid construction approach given by \cite{burt1983laplacian} to avoid quantization errors and ensure that the pyramid levels can accurately reconstruct the input signal. We first filter and downsample the input, then upsample it again to match the prior level's temporal sampling rate; this downsampled-then-upsampled signal is finally subtracted from the input video to calculate the Laplacian pyramid level.

\begin{figure*}
    \centering
    \includegraphics[width=\textwidth]{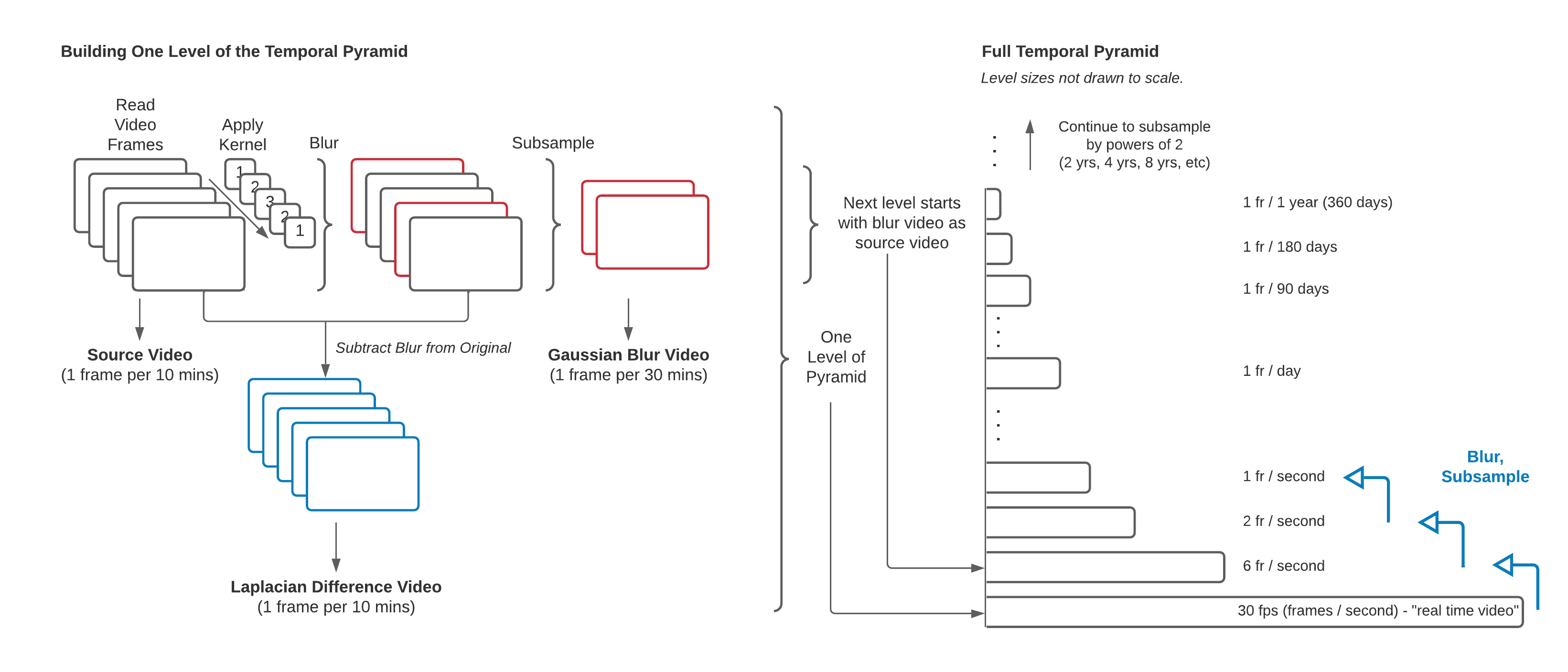}
    \caption{An overview of the temporal pyramid construction process. \textbf{Left:} an illustration of the algorithm for computing one level of the Gaussian and Laplacian pyramid. A source video (the input, or a prior level of the Gaussian pyramid) is blurred in the temporal dimension. The subsequent level of the Gaussian pyramid is computed by subsampling this blurred video, while the Laplacian pyramid level is computed by subtracting the blurred video from the source. \textbf{Right:} The resulting pyramids are the collection of videos generated using the above algorithm.}
    \label{fig:temp_pyr}
\end{figure*}

\begin{figure}
    \centering
    \includegraphics[width=\linewidth]{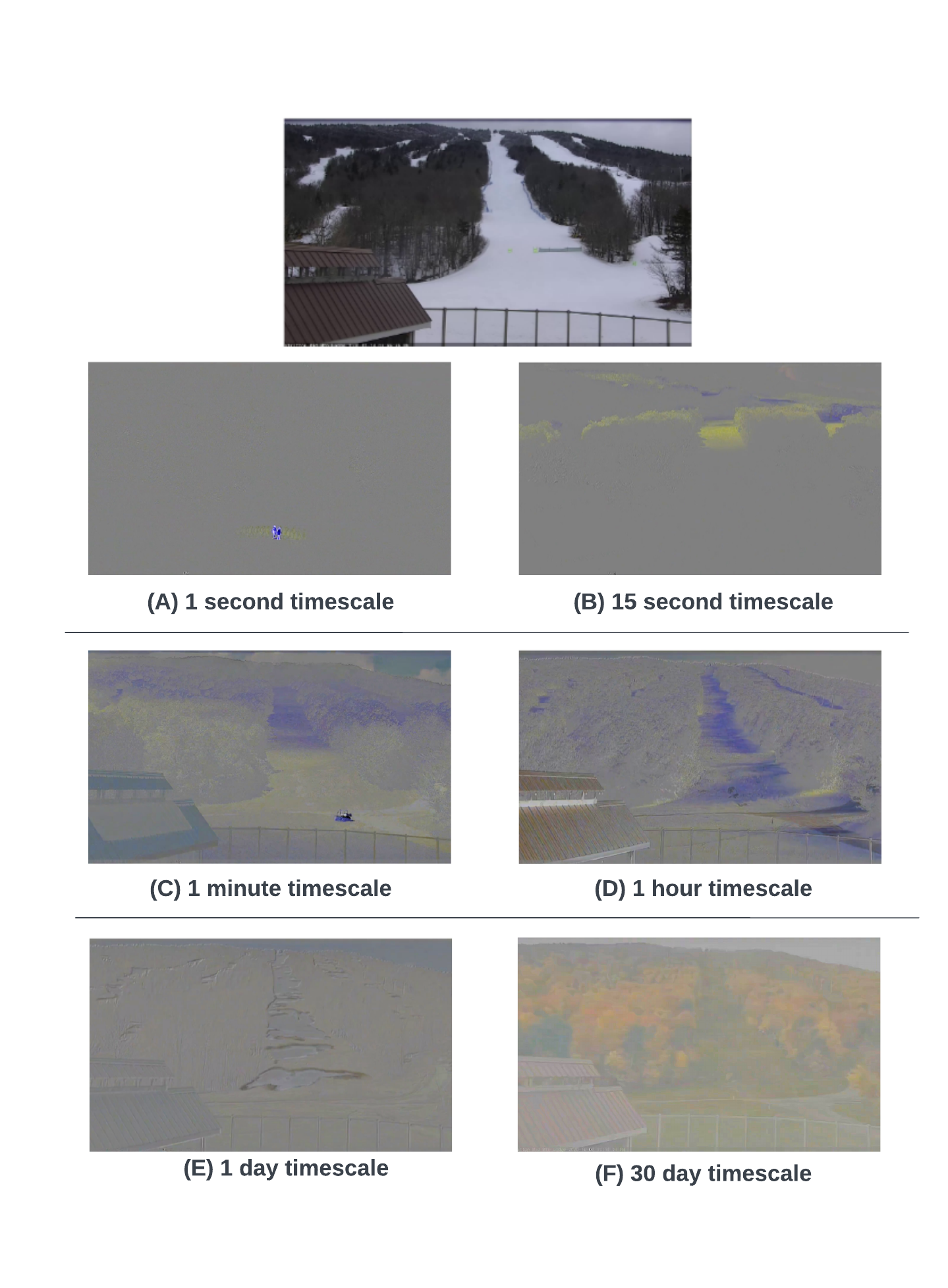}
    \caption{Examples of frames from Laplacian Difference videos at different frequencies, taken from the same webcam but not necessarily from the same days. These show the pixels that changed during that time span. (A) people walking; (B) evidence that the sun peeked out from behind some clouds; (C) a golf cart or similar slow-moving vehicle; (D) evidence of the sun's movement across the sky which has cast shadows of the trees on the ground; (E) outlines of snow patches which likely means that those patches melted over the course of the day; (F) most elements of the scene are visible, including fall colors, which likely means that the autumnal seasonal changes that month affected most pixel values.}
    \label{fig:laplacian_frames}
\end{figure}

The pyramid levels are constructed recursively as shown in \autoref{fig:temp_pyr} and Algorithm \autoref{alg:makepyr}. We begin with a long input video, which serves as the first level of the Gaussian temporal pyramid. Each subsequent level is computed in one pass through the prior level's video, calculating blurred frames from a sliding window of prior level frames. The resulting pyramid levels are themselves videos of the same spatial resolution and covering the same real-world duration in time, but with a reduced frame count. For this reason, each level of a Gaussian temporal pyramid is similar to a timelapse video with a particular sampling rate. One key difference is that blurring across time before subsampling causes short-term motions to blur out in higher levels of the pyramid, thus eliminating aliased content that would appear in a true timelapse video. The levels of the Laplacian temporal pyramid are less intuitive to watch, as they contain only specific bands of temporal frequency content. Sample Laplacian pyramid frames are shown in \autoref{fig:laplacian_frames}.

\begin{algorithm}
\caption{Construct temporal pyramid}
\label{alg:makepyr}
\begin{algorithmic}
\Require $V$, an input video of size $(H \times W \times C \times $ Frames$)$
\Ensure $G = G_{1\ldots N}$, the Gaussian pyramid
\Ensure $L = L_{1 \ldots N}$, the Laplacian pyramid
\For{$i \gets 1 \ldots N$}
    \State $F \gets filter(V)$ \Comment apply linear 1D blur in time
    \State $V' \gets subsample(F)$ \Comment \textit{e.g.}, if stride=3, keep every 3rd frame
    \State $G_i \gets V'$
    \State $F^* \gets filter(upsample(V'))$ \Comment account for quantization error 
    \State $L_i \gets V - F^*$
    \State $V \gets V'$  \Comment set up input for next level
\EndFor
\end{algorithmic}
\end{algorithm}

In a Laplacian temporal pyramid the original source video has been decomposed into multiple component videos. These components can be reassembled to create an exact copy of the source by performing the pyramid-creation steps in reverse order (see Algorithm \autoref{alg:reconstruct}). It is also possible to proceed with this reconstruction while leaving out certain Laplacian pyramid levels (specified in Algorithm \autoref{alg:reconstruct} using the $W$ vector). Reconstructing without the last few Laplacian layers yields a smooth but slower-moving version a Gaussian blur level. We found this smooth temporal upsampling option to be quite useful for some of our videos, as it provides a way to slow down the action so more information can be absorbed by the viewer.

\begin{algorithm}
\caption{Reconstruct pyramid level videos or upsample}
\label{alg:reconstruct}
\begin{algorithmic}
\Require $k \in \{0 \dots N-1 \}$, chosen ending level 
\Require $L_{k+1 \ldots N}$, Laplacian temporal pyramid videos
\Require $G_N$, the top level Gaussian temporal pyramid video
\Require $W \in \{0,1\}^N$, indicator vector of detail levels to reconstruct
\Ensure $R_k$, reconstructed video from level $k$
\State $B \gets G_N$
\For{$i \gets N \ldots k+1$}
    \State $U \gets upsample(B)$ \Comment \textit{e.g.}, if stride=3, repeat each frame 3 times
    \State $F \gets filter(U)$ \Comment use same filter from pyramid construction
    \State $B \gets F + (L_i \times W_i)$ \Comment add detail layer if applicable
\EndFor
\State $R_k \gets B$ \Comment reconstructs original when $k=0$ and $W_{k+1 \dots N}=1$
\end{algorithmic}
\end{algorithm}

Each level of the pyramid represents a specific \textit{timescale}. For instance, the base frame rate of most webcam video is 30 frames per second (fps), or 1/30th of a second per frame. Motions easily visible at this frame rate can be thought of as belonging in the ``1/30th of a second'' timescale. Higher levels of the pyramid represent changes happening at a frequency of ``once per second'' or ``once per day'' or even ``once per month.'' 

\subsection{Adaptations for Months of Static-Camera Video}

The prior section described a straightforward generalization of the Gaussian and Laplacian image pyramids to construction of temporal pyramids. We now describe a few simple adjustments we made to adapt these pyramids for the use case of visualizing and exploring long, fixed-camera video streams.

\subsubsection{Variable Downsampling Rates}
In image pyramids, the filter width and subsampling rate are parameters that are traditionally tuned according to the application. For example, to achieve scale invariance for computer vision algorithms, a subsampling rate smaller than 2 is often desirable to detect objects at a densely sampled range of sizes. In our application, the pyramid is unlikely to miss anything, even with a larger sampling rate because motions and dynamics tend to be visible in a range of timescales. For example, in a 30 frames-per-second (30fps) video of a person walking through a courtyard, the person might take 6 seconds to walk through the scene, and thus would appear, moving increasingly quickly and increasingly blurred, in at least the first four or five levels of the pyramid.

While we initially simply used a downsampling rate of 2, problems arise when the temporal sampling rate of each pyramid level is not aligned with intuitive units of time. As discussed in \cite{aigner2011timevis}, modeling time has many complexities to consider. For example, if our input video (Gaussian pyramid level 0) is captured at 1/30 second per frame (30 frames per second) and we chose a fixed scale factor of 2x, then level 5 of the pyramid would cover 1.066 seconds per frame, level 10 would cover 34.133 seconds per frame, and level 22 would cover 1.618 days per frame. In addition to being less intuitive for interpretation, these sampling rates can introduce aliasing at higher sampling rates due to periodic phenomena such as the day/night cycle or seasonal changes. Since our goal is to visualize and discover structure at long time-scales, it is important to have sampling rates lined up with known patterns such the 24-hour cycle of the day and the 365-day cycle of the year.  Achieving this alignment requires choosing different subsampling rates for different levels of the pyramid. See Supplementary Material for details of the sampling rates used and timescales represented at all levels of our pyramids. We used strides of 2, 3, and 5, which necessitated the use of different blur filters depending on stride. The one-dimensional blur filter applied across frames was [1,2,2,1] when the stride was 2; [1,2,3,2,1] when the stride was 3; and [1,2,3,4,5,4,3,2,1] when the stride was 5.

\subsubsection{Scaling to Months and Years of Video}
The algorithm as described thus far requires a full pass through each pyramid level to compute the next; because the layers sizes shrink exponentially, this requires the equivalent of roughly 2 passes through the full input video, for an $O(n)$ runtime. However, for months-long input videos such as ours, this is still very slow and can be easily parallelized. To accelerate the computation of pyramids, we compute one-day pyramids in parallel on a cluster, then merge the one-day pyramids to compute the higher pyramid levels. These one-day pyramids are computed up through level 15, where the Gaussian blur video for the entire day is 1 frame, and the corresponding Laplacian pyramid level shows activity changing on a 12-hour timescale. The one-day pyramids are then merged by stitching together each day's 1-frame Gaussian `video' into a full blur video for level 15, which is then used as the source video for the construction of the remaining pyramid levels. 

We also parallelize the creation of years-long videos in the same manner, running each year separately and then stitching them together and continuing the construction of multi-year pyramid levels. In addition to being efficient, this also allows us to approximate a 365- (or 366-) day year with only 360 frames, which is necessary in order to use only sub-sampling rates of 2, 3, and 5. For each individual year, after analyzing which days have the most missing frames, we choose 5 days to remove from the pyramid (or 6 days for a leap year). If we used this 360-day year and did not compute each year separately, we would end up with a true 360-day timescale and some aliasing over the course of multiple years, where the year shown would slowly get out of alignment with the calendar year. When we parallelize the years, we end up with 1 frame per year at level 21 (the 1-year timescale), which line up with calendar years, and the higher levels can be built on that solid foundation. We currently sub-sample with powers of 2 for multiple-year timescales, but it would be possible to sample by 2 and then 5 (or 5 and then 2) in order to create a 1-decade timescale.

Missing data is filled in with all-black frames in our temporal pyramid videos. See the supplementary material for a more detailed description of how we handled missing data.

%% file: sections/4_spectrogram.tex
\section{Video Spectrograms\label{sec:spectrograms}}

\begin{figure}
    \centering
    \includegraphics[width=\linewidth]{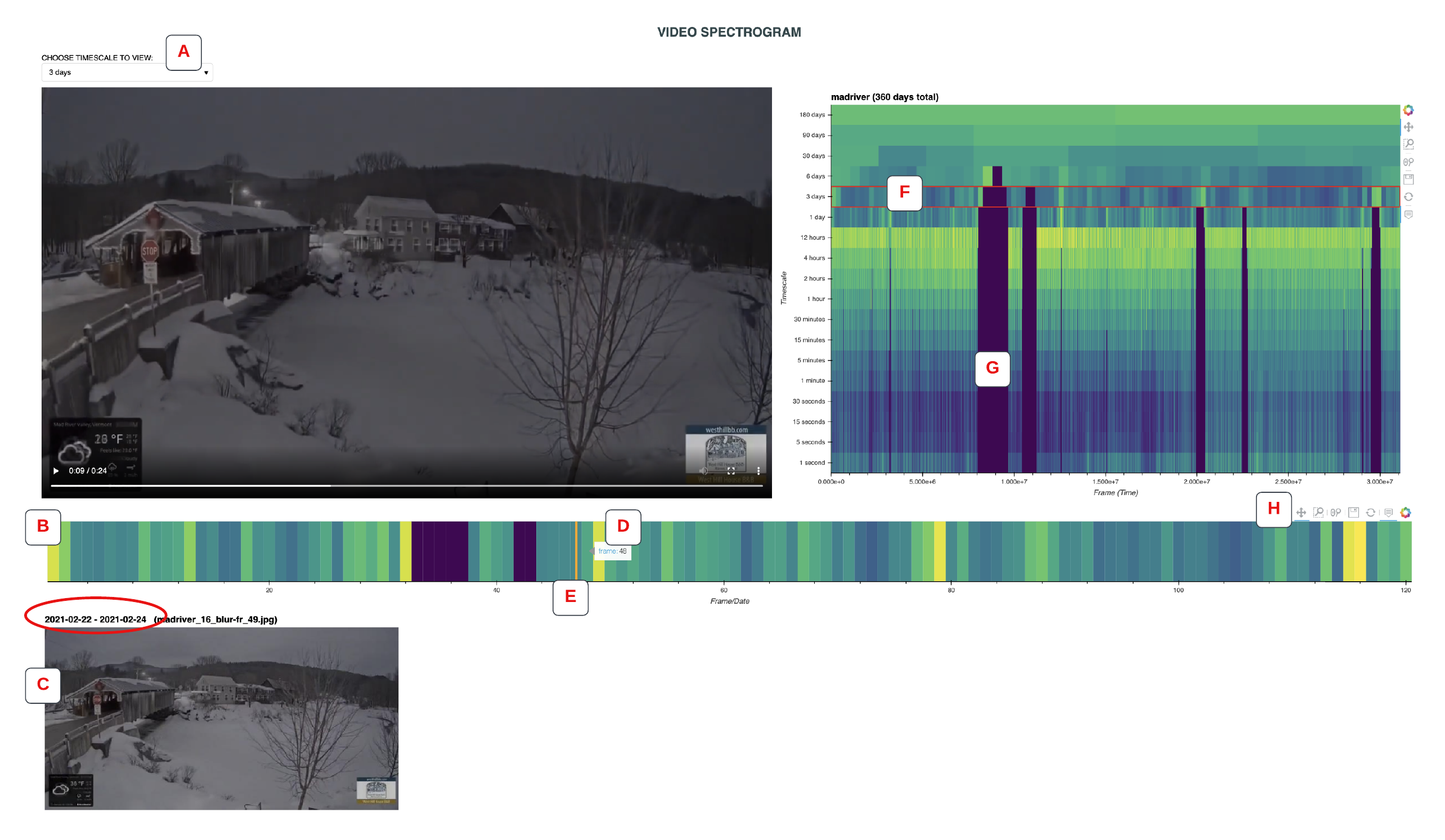}
    \caption{Drop-down menu (A) for choosing level to view. Video player updates with appropriate level video. Chosen level is highlighted on the full spectrogram (F) as well as enlarged below the video (B). On mouse hover (D) a thumbnail image of that frame shows below (C). As the video plays, a vertical orange line (E) will travel along the single-level spectrogram plot, aligning the date/time between video and plot. Areas with missing data, such as (G) show up clearly. Toolbars (H) allow for zoom and save.}
    \label{fig:full_spectro}
\end{figure}

At this point, we have described an approach for parsing out temporal frequency content of a long video stream into timescales by constructing a temporal Laplacian pyramid.  Watching just the upper level videos of the pyramid is an efficient way to gain an overview of temporal dynamics and long-term events because they are short but distill important information. However, the pyramid itself does not make it any more tractable to watch the entirety of the lower levels, which still have very long durations. To help address this, we propose a visualization tool called the Video Spectrogram that facilitates interactive navigation and exploration of the pyramid levels.

The Video Spectrogram user interface evolved to include multiple elements, as shown in \autoref{fig:full_spectro}; however, the main element and key idea is a 2-dimensional plot that provides an overview of the entire pyramid by showing time on the horizontal axis and timescale (frequency) on the vertical axis. Each cell in this time/frequency grid represents a 2D frame from one of the pyramid videos, which would be unwieldy to visualize in such a small space; instead, we abstract the spatial details and display a single quantity that captures aggregate activity.

By construction, the Laplacian pyramid layers are ``difference'' frames representing only content that has changed at the corresponding timescale. Therefore, the aggregate activity for a given frame in a timescale can be measured by taking a norm of the Laplacian frame. We chose the $L^2$ norm (i.e., the square root of the sum of squared pixel values), and display it on a logarithmic scale. We experimented with other norms ($L^1$) and color scales (linear). Because we aggregate across pixels, the $L^2$ norm gives more weight to spatially smaller changes with larger magnitude versus more widespread, smaller-magnitude changes. The logarithmic color map does a better job of showing contrast in low-activity regions, allowing subtler patterns to be detected when overall activity levels are low.

We compute the norm for each frame in each Laplacian pyramid level, and display the resulting values as a 2-dimensional heatmap as shown in \autoref{fig:teaser} and \autoref{fig:full_spectro}, where each tile in the heatmap is the norm of one Laplacian pyramid frame. Tiles in higher timescales become wider because the same temporal extent is represented using fewer frames at higher pyramid levels.

The temporal pyramid videos and the spectrogram plot are closely linked; the purpose of the spectrogram is to help explore the pyramid, so we include a large and prominent video player to show the pyramid videos. The full-spectrogram plot is good for an overview; however, we found that since we generally watch the video from one level at a time, it was useful to enlarge the portion of the heatmap corresponding to the level being watched in the video. We visualize this single-level spectrogram below the video, along with a moving vertical line that travels along the plot as the video plays. As the user sees an event unfold in the video they can get a sense of what the spectrogram shows during the event. The full-spectrogram plot always includes a red outline marking the level and/or date being viewed in the current video, for a `you are here' connection to the bigger picture. Both plots have pan and zoom functionality to assist with overview-to-detail visualization.

In order to go in the other direction, and see what the video content is like at a particular spot by starting from the spectrogram, we implemented a mouse-over functionality whereby a thumbnail image of the corresponding video frame shows up underneath the plot when the mouse hovers over a cell of the single-level spectrogram. Easy access to those thumbnail images gives the user hints about the reason for the structure in the heat map and helps determine whether it's useful to drill down or investigate further in that area. The thumbnail-on-hover functionality also makes it possible to ``scrub'' through the video for that timescale by dragging the mouse horizontally over the plot at any speed.
 
Users can navigate to different levels of the pyramid using a drop-down menu at the top of the user interface. At the 5-minute timescale or lower, the user is given the choice to view a particular date instead of the whole timespan, since those levels are very large and the assumption is that only a portion will be watched. Once a date is selected, it will stay selected while the user navigates down to lower levels, making it easier to `drill down' on interesting content. Also, the user can stay on one level and easily select a different day on that same level to view, which helps for comparing dynamics across different days. When a user hovers to view a thumbnail, the image also displays the date and time (or timespan) represented by that frame. This information is valuable for following the thread of an event from upper to lower levels to gain greater detail and pinpoint the timing of that event. Quick access to date and time information also allows the user to make use of their own knowledge of specific past events at that location or patterns of life relevant to that scene.

%% file: sections/5_results.tex
\section{Results}

\begin{figure}
    \centering
    \includegraphics[width=\linewidth]{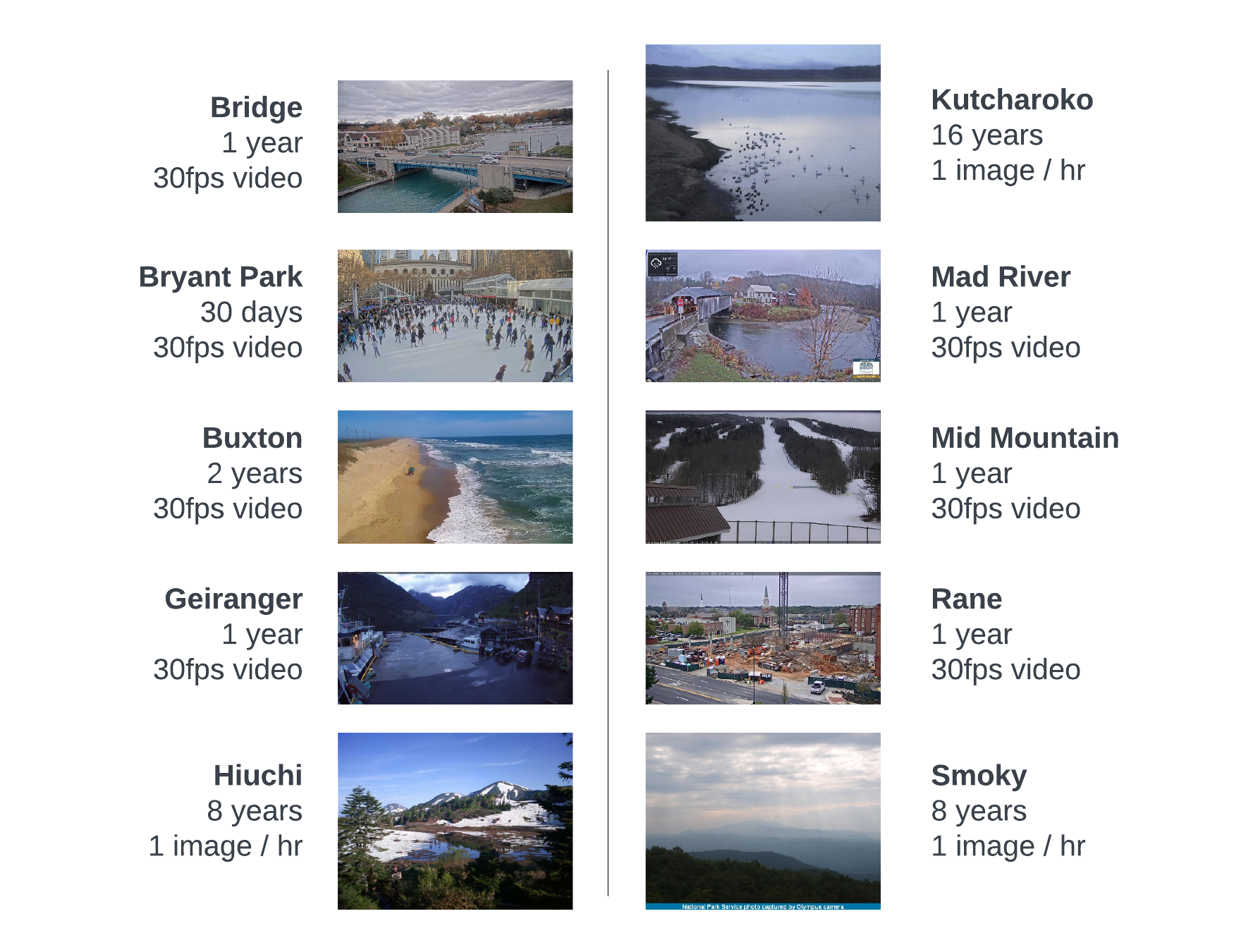}
    \caption{Sample frames from our datasets, along with the names we are using to refer to them in this paper, their covered timespan, and their base frame rate. More details in supplemental material.}
    \label{fig:dataset_list}
\end{figure}

To demonstrate our approach, we scraped or downloaded 10 datasets captured by outdoor webcams, with lengths ranging from 30 days to 16 years. Details for these datasets are included in the supplemental material, and a quick visual reference is included in \autoref{fig:dataset_list}. The pyramid videos and spectrograms revealed interesting dynamics and structure at a range of timescales. Below are some general observations, as well as specific findings for several datasets.

\subsection{Cycles and Visualization of Periodicity}

Events that happen repeatedly stand out with clarity in our pyramid videos. The day/night cycling is the obvious example, and this is noticeable in all of our datasets. However, we found many other examples of cyclic activity that appeared in the videos as distinctive repetitive patterns. Tidal patterns were apparent in the Buxton oceanside dataset as well as the Geiranger ferry dock dataset. Geiranger shows boats rising and falling next to a dock. In the timescales below 1-day, the boats move up and down as the video progresses, and in the timescales longer than 1-day the movement becomes averaged and the video shows a ghostly blur encompassing all of the vertical positions of the boat over time. The cyclic nature of seasons becomes obvious in the datasets with multiple years, such as Hiuchi, Kutcharoko, and Smoky, where we can view the seasons changing fast enough that we understand the similarities of the cycle from year to year. The switching between white snowy winter and green leafy summer becomes a visual rhythmic pulse at the higher pyramid levels, just like the day/night changes at the lower pyramid levels.

On a smaller scale, the natural cycles of plant growth are nicely visualized in the Mad River dataset. This scene includes a deciduous tree in the foreground, as well as bushes and other trees on the edge of a river. The tree in the foreground loses its leaves and grows them again, and the bushes and other plants can be viewed getting larger in the summer and smaller in the winter. Another very interesting discovery is how the branches of the foreground tree droop at night and perk up during the day, which becomes more noticeable because it happens repetitively. This dataset seems to have its camera recording with infrared at night, which fortunately makes the tree always visible.

In addition to the pyramid videos themselves, the Video Spectrogram also seems to be especially good at visualizing cyclic events. The day/night cycle is very obvious at the 12-hour timescale, with (usually) more activity and a lighter color on the spectrogram for the half of the day which is mostly daylight, and (usually) a darker color for the less-active night. The dark and light colors on the spectrogram switch back and forth creating a distinctive pattern at that level of the pyramid for most of our datasets. In the higher levels of the multiple-year datasets, the seasonal pulsing is also clearly visualized with the spectrogram colors.

\begin{figure}
    \centering
    \includegraphics[width=\linewidth]{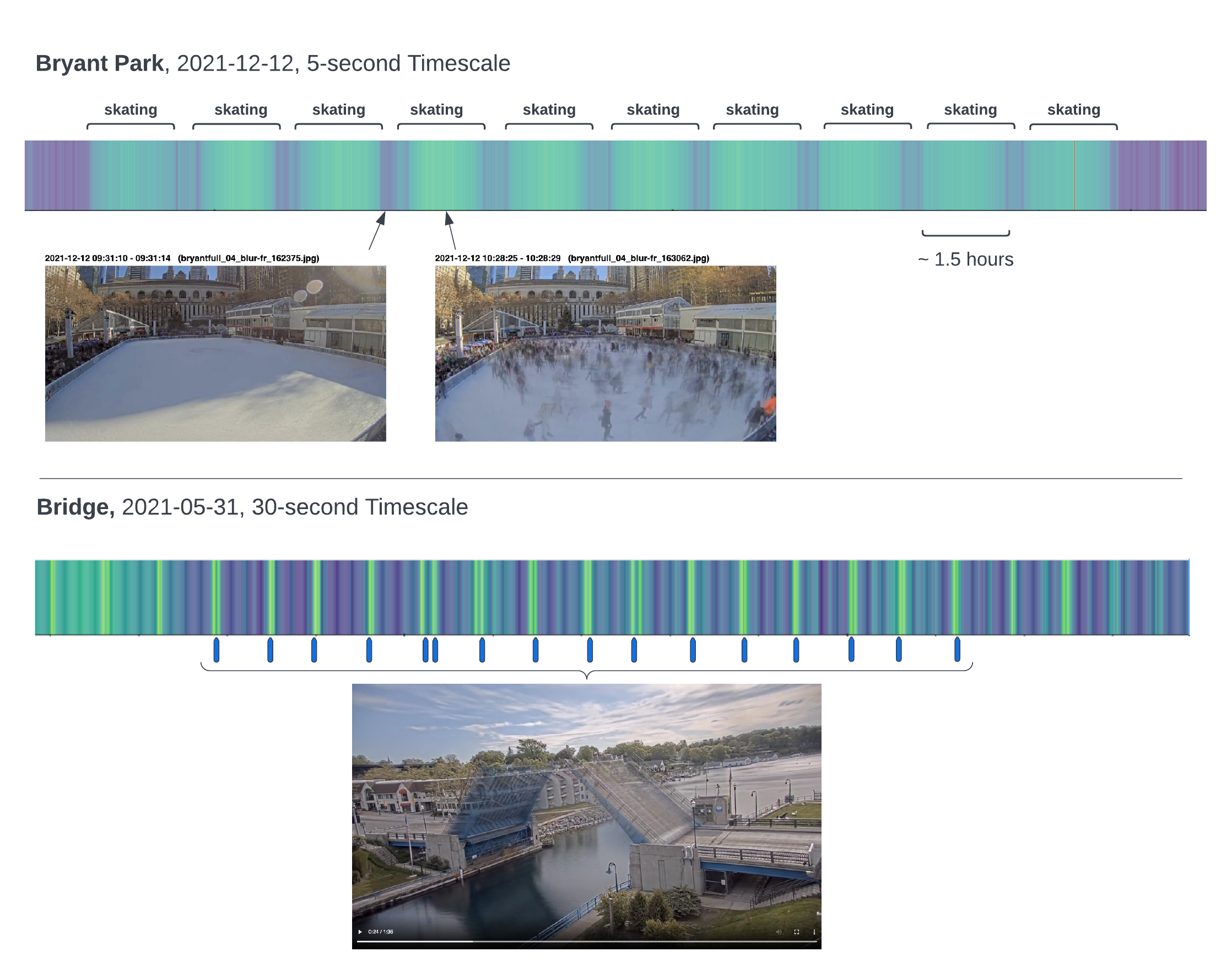}
    \caption{Examples of periodic human activity showing in the spectrogram.}
    \label{fig:bryantbridge}
\end{figure}

We also found other examples where the periodicity of human activity showed up clearly in the spectrogram (\autoref{fig:bryantbridge}). In the Bryant Park ice rink dataset, the spectrogram had lighter colors during the times the rink was busy with skaters and darker colors for the times when the ice was cleared. This seemed to happen on a cycle of about an hour and a half, presumably a planned timing. Another example is from the Bridge dataset, where the spectrogram shows a light bar every time the drawbridge goes up which happened frequently on Memorial Day in 2021. Since this was a repeating event over the course of that day, its pattern on the spectrogram was more noticeable than it was on the days where the drawbridge only went up once or twice.

\subsection{Multiscale Visualization and Drill-Down Navigation}

\begin{figure}
    \centering
    \includegraphics[width=\linewidth]{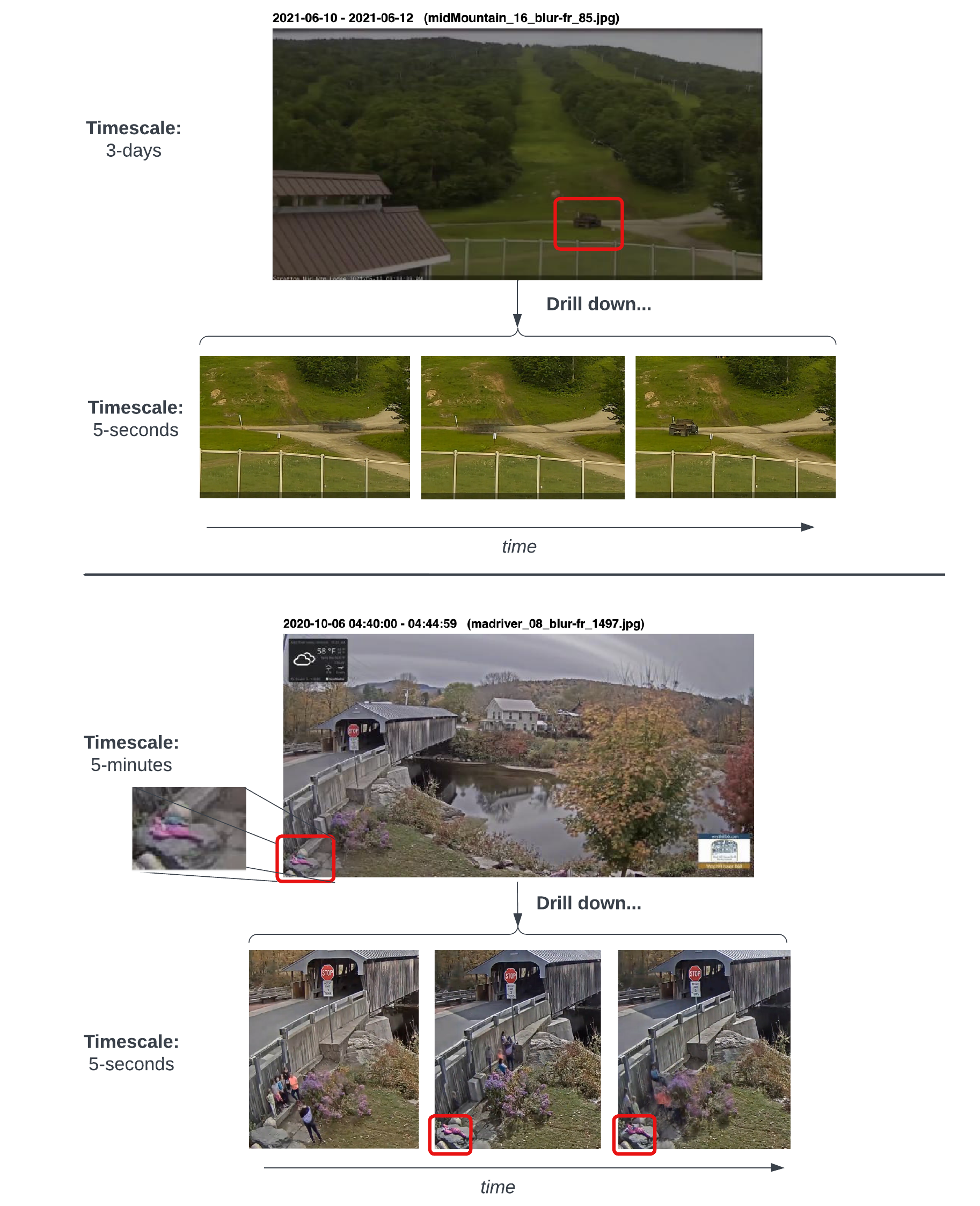}
    \caption{TOP: In Mid Mountain, a truck abruptly appears and quickly disappears in the 3-day timescale. Drilling down to the 5-second level is necessary to learn that the truck drives in from the left and backs into its parking spot. BOTTOM: In Mad River, a pink jacket left on a rock shows up briefly at the 5-minute timescale. Drilling down reveals a group of people moving around.}
    \label{fig:drilldown}
\end{figure}

The video spectrogram tool connects events at different timescales by virtue of using a common timeline. If the viewer sees a short blip of some interesting or anomalous event at a higher timescale, the user allows for pinpointing the general date/time of the anomaly and drilling down to lower levels at that same time in order to see more detail. For example, one might see a truck appear `out of nowhere' in the 6-day timescale, then quickly drill down to the day it appeared and view the 5-second timescale on that day in order to see which direction the truck drove in from before it parked (\autoref{fig:drilldown}, top).

Most real-world events do not fall neatly into one discrete timescale, and this often means that an interesting event can first be discovered from the bird's-eye perspective of a higher pyramid level and then the full extent of the occurrence can be discovered and viewed by drilling down to lower levels. We found an example of this in the Mad River dataset (\autoref{fig:drilldown}, bottom), when a bright pink object catches the eye briefly and then disappears during the 5-minute timescale on October 6, 2020, in a corner of the scene. Drilling down to the 5-minute level, the pink object appears to be a pink jacket left on a rock but it still disappears quickly. Drilling down to the 1-minute level we can see fast-moving people and we also see that the pink jacket moves from one rock to another. However, it is not until we drill down to the 5-second timescale that we can make out the group of people and their general movements. The fact that they left a bright pink jacket in one place for about 15 minutes, while they themselves moved faster, left a clue to their presence in that longer 15-minute timescale video. 

The multiple timescale visualization also provides a useful and possibly educational demonstration of the role and scope of human activity in a particular scene. In the Bryant Park dataset, the lower level pyramid videos show crowds of people skating. However, the higher level pyramid videos show an eerily empty skating rink, with no humans in sight. At the 4-hour timescale and above, it is mainly the lighting changes, and certain infrastructure changes that stand out (such as a rink-side tent being erected for a while). In the Mid Mountain ski slope dataset, tire tracks and ski tracks in the snow provide a clue to human activity at the lower levels of the pyramid, but we don't see the humans making those tracks unless we drill down. The same thing can be seen with tire tracks and footprints in the sandy beach of the Buxton dataset. With both snow and sand, the evidence of human activity is melted or eroded away fairly quickly. In contrast, the Rane construction dataset shows a scene where the long term effect of human activity is exactly the point, and in that case it is definitely instructive to drill down and see exactly which lower-timescale activities were responsible for the higher-timescale view of a building being constructed.

\subsection{Discovery of Anomalies}
\label{sec:anomalies}
The pyramid videos by themselves, as well as the spectrogram tool, can be useful for surfacing anomalous events. Missing data is the most obvious anomaly to find, and sections of missing data show up as all-black frames in the videos and as solid dark colored areas of the spectrogram, as can be seen in \autoref{fig:full_spectro}. Corrupted data is another kind of anomaly that shows up in the videos and the spectrogram. For instance, in the Buxton coastline dataset the high-level videos have a section that shows a static night scene which is clearly out of place since it lasts longer than a single night both in the pyramid video and on the timeline. We traced the problem back to the original footage, where a static image was looped for a while.

Many anomalies are related to the camera itself, the most common of which is a sudden change in camera angle or placement. The Mid Mountain dataset includes a few months during ski season where the camera zooms or moves closer to the ski slope. The Buxton beach video includes a section where the camera faces out to the ocean instead of along the coastline. Sometimes camera errors can be seen, such as a day in the Hiuchi dataset which included camera footage of an office ceiling when normally the scene is outdoors in the mountains. That event occurred directly after a long period of missing data, so we suspect the camera was being repaired before being reinstalled outdoors. When there is a rainstorm or snowstorm, the video will often show raindrops on the camera lens for a short while at the lower timescales. The most entertaining camera-related anomalies we found occured when birds perch in front of the camera (Buxton) or a spider builds a web on it (Mad River). 

\begin{figure}
    \centering
    \includegraphics[width=\linewidth]{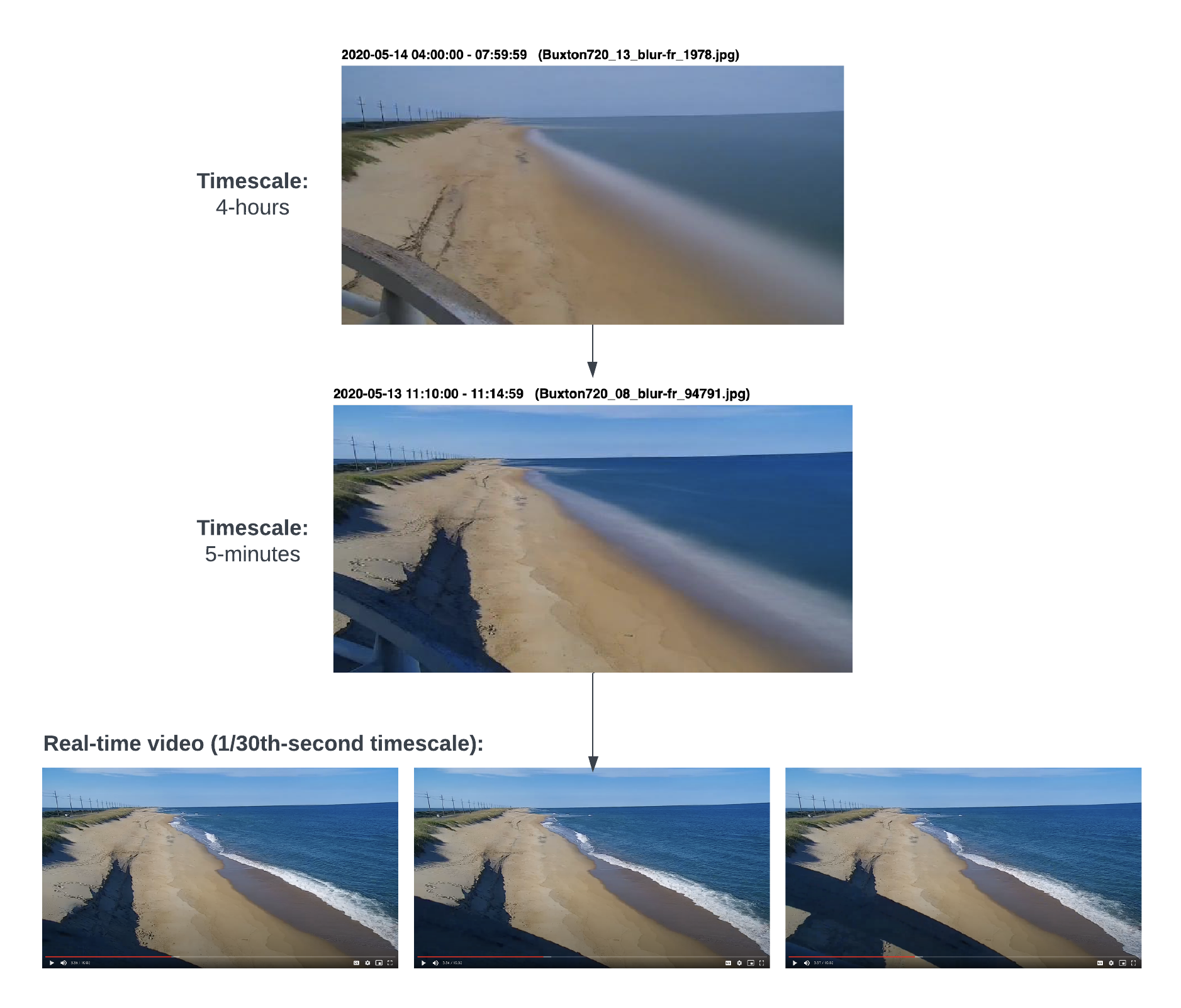}
    \caption{In the Buxton dataset, at the 4-hour timescale, a railing quickly and obviously appears directly under the camera. Drilling down, we had to go to the original real-time recording to see it being put into place.}
    \label{fig:buxton_railing}
\end{figure}

In contrast with standard timelapse, the pyramid videos are visually less cluttered and they can be upsampled to adjust the rate of change to be easier to absorb. Anomalous events are more distinctive against this backdrop and are thus fairly easy to spot. For instance, when watching the Buxton dataset video for the 4-hour timescale, we noticed that a railing suddenly appears directly underneath the camera. At that level we could localize it to May 14, 2020, using the Video Spectrogram. We went directly to the 5-minute timescale for May 14, 2020, but the railing was there at the beginning of the day, so we switched to May 13, 2020, and drilled down further. We could see the railing put into place in the real-time original video for May 13, 2020. Even then it was very fast and rather anti-climactic since there was no visual of the person putting it in place (see \autoref{fig:buxton_railing}).
 
\subsection{Long-term Dynamics and Understanding}
Our pyramid videos provide a window into the reality of long-term dynamics without sacrificing much verisimilitude. There is some level of blurring that occurs at the upper levels of the pyramid, since it has averaged a lot of small changes over time. We also see the blending of day and night scenes. These factors mean we sacrifice some precision and spatial resolution at the longer timescales; however, we found we were still able to discern larger patterns. For instance, in our Mid Mountain dataset, the gradual pattern of snow melt on a ski slope over the course of days or weeks stands out with clarity at the 6-day timescale. In our Buxton oceanside dataset, at the 1-month timescale, the water's edge can be seen slowly changing its position relative to the beach, slowly rising and retreating much slower than the tides.

Another way these pyramid videos contribute to understanding of long-term dynamics is through the knowledge that anything showing up in a particular timescale must have generally stayed in the same place for a long enough time, related to the timescale. In the 1-hour timescale, a car driving by would not show up but a car parked in one spot for at least an hour would show up, and the duration of its appearance in the video would correspond to how long it stayed parked.

\subsection{Direct Comparison with Timelapse}
    For a few datasets, we constructed standard timelapse videos by subsampling at different rates and compiling the resulting frames together into videos for each timescale. At the top levels of the pyramid this resulted in extremely short timelapse videos (less than one second), which were thus not very informative or interesting. However, going down the pyramid levels, once the videos were at least a few seconds long they did provide a good baseline for comparison with our pyramid videos. We found consistent results among all of the datasets which are summarized below.
    
    \subsubsection{High Levels: 1-day Timescale and Above}
        The visual smoothness of our pyramid videos stands in stark contrast with the timelapse videos. At the higher levels especially, each frame of the timelapse is far removed in time from its neighboring frames, increasing the likelihood of major discontinuities in lighting, weather, and other large scene elements. The timelapse videos show the viewer all of these images in rapid succession and the effect is visually chaotic. Only the most obvious changes can get absorbed by the viewer. The rest of the changes are likely to get lost in the noise.

        Also, smoothly upsampling our videos to spread out over a longer duration makes them more informative and watchable than the timelapse videos, even after using the video player tools to slow the timelapse videos down to quarter speed. At the 3-day timescale, the timelapse video duration was 4 seconds. Slowing it down to quarter time extended it to 16 seconds. However, our upsampled pyramid video for that timescale was 24 seconds.
        
    \subsubsection{Middle Levels: 2-hour to 12-hour Timescales}
        The timelapse videos from these levels are almost unwatchable because of the strobe effect caused by rapid switching between day and night as the video progresses. This problem would likely be fixed by the complete removal of night-time frames. We did not test that idea, but we believe that even with night frames removed, the timelapse videos from these levels would still suffer from similar faults as they do in the higher and lower levels. Also, for a fair comparison we would also need to remove the night frames from our pyramid videos, and this would likely improve the watchability of those videos as well. We have currently bypassed the strobe effect problem in our pyramid videos by upsampling them so they take longer to watch while the pulsing from day to night happens at a gentler cadence. If we removed night frames, we would not need to upsample so much and could watch shorter videos. However, for the majority of our datasets there is interesting visual activity during the night hours which we would not want to arbitrarily excise from the video for the sake of watchability or efficiency. 

    \subsubsection{Low Levels: 1-hour Timescale and Below}
        The timelapse videos are more watchable and informative at lower levels than they are at the higher levels. When comparing a timelapse video with a non-upsampled pyramid video (i.e., of the same length), the pyramid video provides only a slightly better viewing experience because of its smoothness. Upsampling our pyramid video definitely improves its viewing quality.
        
        The most interesting comparison occurs at the very lowest levels (1-minute timescale and below), where we can watch videos for one day at a time and see fast-moving activity such as people skiing and cars driving. In the timelapse videos, fast moving people and cars end up aliased, meaning they show up as a completely solid object and disappear quickly, without a clear trajectory. By contrast, in the pyramid videos, fast moving people and cars will show up as a line of ghostly versions of themselves, along their trajectory. They will only solidify if they stay in one place for long enough. This provides the viewer with more information than the timelapse videos provide. As an example, in the Rane construction site dataset car traffic can be seen on the road in front of the building site at the 15-second timescale. In the timelapse video, we see cars at night and during the day and we can get a general sense that there is less traffic at night. However, in the pyramid video the day/night traffic difference is clearer. We can barely register a blur for night-time traffic, and there is clearly more traffic during the day. It is blurry for the most part, except for when cars stop at the stoplight at regular intervals, at which point they `solidify' into a clear line of cars. There is a rhythm to this blurred/not-blurred traffic, presumably corresponding with the traffic light schedule. Also, when the cars are stopped we can see that there are almost always more cars in the right lane (possibly getting ready to turn right) This is more information than we would ever glean from the timelapse videos and provides a sound argument for the very basis of our video temporal pyramid.

%% file: sections/6_discussion.tex
\section{Discussion, Limitations, and Future Work}

This work proposes a novel way to visualize the passage of time and explore videos that are too large to be practically explored using traditional tools. We also specifically address phenomena that occur at much longer timescales than most existing methods; these phenomena are present and interesting in our application due to the extreme duration of our videos. Our method compares favorably to naive timelapse videos. Even as an imperfect visualization tool, timelapse videos have been put to good use in a diverse range of applications, such as construction site monitoring \cite{tibaut2018construction, yang2015construction}, environmental monitoring \cite{liu2016glacier, hartill2020fishing, seyednasrollah2019phenology}, art \cite{hansen2018art}, education \cite{vollmer2018education, nakamura2019education}, ecological awareness \cite{buckley2017ecology, sierraclub2014}, and more. An improvement to existing timelapse techniques could benefit all of these existing applications and possibly lead to interesting new applications. It might also help facilitate a shift in perspective towards long-term thinking. Humanity's short-term or `real time' bias can make it difficult to tackle long-timescale issues like climate change or urban sprawl. If we don't \textit{see} it happening, we don't care about it as much. Visualization tools can help us \textit{see} it \cite{nakamura2019education, monea2021education}.

\subsection{Limitations}
Our method has some important limitations. One is the requirement that the camera viewpoint be fixed. This ensures that changes in the video are due to the scene, rather than camera motion; however, if camera angle changes are infrequent then the spectrogram is minimally affected, and in fact our method is useful for discovering unusual camera events such as a change in viewpoint as discussed in \autoref{sec:anomalies}. 

Another limitation is that the top two or three levels of the pyramid are usually not very informative because there are not enough frames available for any changes to register when stitching the frames together. In a pyramid built from one year's worth of data, the 90-day timescale will only have 4 representative frames. However, in a pyramid built from 8 year's worth of data, the 90-day timescale will include 32 representative frames. We can discern changes over 32 frames much more easily than over 4 frames. 

We also noticed that as the pyramid levels progress higher, the edges of all scene elements tend to become slightly less sharp with each new level. This is probably caused by very small camera movements that register as brief `whole scene changes' with an effect in the pyramid that compounds as levels are built recursively. This effect could possibly be mitigated by the addition of a video stabilization preprocessing step. Simple feature matching-based image alignment techniques \cite{szeliski2007image} could be used to align the frames to minimize movement due to camera shakiness. A similar feature matching technique could be used to manage camera viewpoint changes as well. We found that the noise present in our datasets was small enough that these techniques were not necessary, but they could be used to boost visual quality if desired. They could also help our method generalize to more datasets, such as those with automated and periodic changes in viewpoint, or fixed-viewpoint cameras that exhibit noticeable motion due to wind.

\subsection{Future Work}
Our temporal pyramid computes a very simple, low-level measure of intensity change from one frame to the next. In the spirit of Viz-a-Vis \cite{romero2008vizavis}, we intend to explore more sophisticated types of analyses that can be aggregated into heatmaps to show more high-level and/or task-specific measures of activity. For example, optical flow could be used to measure motion rather than per-frame intensity change at multiple timescales (similar to \cite{wehrwein2021scene}). In scenes with specific object categories of interest (e.g., people, cars etc.), object detection or crowd counting techniques could extract more meaningful trends which could then be visualized in a similar time-frequency spectrogram. 

Although our method was not designed with the intent of video anomaly detection, it could provide the basis for some new techniques in that area. One of the many existing video anomaly detection methods \cite{nayak2021anomaly} could possibly be applied to upper level pyramid videos in order to quickly and automatically surface unusual events at those timescales, which might yield insight upon drill-down to lower levels. For instance, detecting the origin of an unattended bag, after the fact, would be likely made easier with the aid of a temporal pyramid.

Another direction for future work is to explore different types of visualizations for our spectrogram, other than a heatmap. For example, a circular or radial representation might be useful for visualizing periodic events. We would also like to find ways to more easily compare different days with each other (or different years, months, etc.), even if the days chosen for comparison are far removed from each other in time.

\section{Conclusion}
In this paper, we presented the Video Temporal Pyramid -- a multi-scale lens through which to view the passage of time via a process that distills activity happening at different timescales in long fixed-camera video streams. We also presented the Video Spectrogram, a time-frequency visualization to facilitate exploration and discovery in our pyramids. The pyramid videos present a novel alternative to standard timelapse techniques, providing a smooth viewing experience that allows for the absorption of more information about how a scene changes over time. And the spectrogram visualization is the first example of what we believe is a more general and potentially useful class of time-frequency representations for video visualization.

%% file: ms.bbl
\begin{thebibliography}{10}
\renewcommand*{\sfdefault}{PTSansNarrow-TLF}

\bibitem{aigner2011timevis}
\href{https://doi.org/https://doi.org/10.1007/978-0-85729-079-3}{W.~Aigner,
  S.~Miksch, H.~Schumann, and C.~Tominski}.
\newblock \href{https://doi.org/https://doi.org/10.1007/978-0-85729-079-3}{{\em
  Visualization of time-oriented data}}.
\newblock
  \href{https://doi.org/https://doi.org/10.1007/978-0-85729-079-3}{Human-computer
  interaction series}.
  \href{https://doi.org/https://doi.org/10.1007/978-0-85729-079-3}{Springer},
  \href{https://doi.org/https://doi.org/10.1007/978-0-85729-079-3}{2011}.
  \href{https://doi.org/10.1007/978-0-85729-079-3}
{doi: \textsf{%
10\hspace{.1pt}\discretionary{.}{%
}{.}\hspace{.4pt}1007\discretionary{/}{%
}{/}978\discretionary{%
}{-}{-}0\discretionary{%
}{-}{-}85729\discretionary{%
}{-}{-}079\discretionary{%
}{-}{-}3}}


\bibitem{ali2019timecluster}
\href{https://doi.org/10.1007/s00371-019-01673-y}{M.~Ali, M.~Jones, X.~Xie, and
  E.~M. Williams}.
\newblock \href{https://doi.org/10.1007/s00371-019-01673-y}{Timecluster:
  dimension reduction applied to temporal data for visual analytics}.
\newblock \href{https://doi.org/10.1007/s00371-019-01673-y}{{\em The Visual
  Computer}}, \href{https://doi.org/10.1007/s00371-019-01673-y}{35:1--14},
  \href{https://doi.org/10.1007/s00371-019-01673-y}{06 2019}.
  \href{https://doi.org/10.1007/s00371-019-01673-y}
{doi: \textsf{%
10\hspace{.1pt}\discretionary{.}{%
}{.}\hspace{.4pt}1007\discretionary{/}{%
}{/}s00371\discretionary{%
}{-}{-}019\discretionary{%
}{-}{-}01673\discretionary{%
}{-}{-}y}}


\bibitem{barnes2010tapestries}
\href{https://doi.org/10.1145/1778765.1778826}{C.~Barnes, D.~B. Goldman,
  E.~Shechtman, and A.~Finkelstein}.
\newblock \href{https://doi.org/10.1145/1778765.1778826}{Video tapestries with
  continuous temporal zoom}.
\newblock \href{https://doi.org/10.1145/1778765.1778826}{{\em ACM Trans.
  Graph.}}, \href{https://doi.org/10.1145/1778765.1778826}{29(4)},
  \href{https://doi.org/10.1145/1778765.1778826}{July 2010}.
  \href{https://doi.org/10.1145/1778765.1778826}
{doi: \textsf{%
10\hspace{.1pt}\discretionary{.}{%
}{.}\hspace{.4pt}1145\discretionary{/}{%
}{/}1778765\hspace{.1pt}\discretionary{.}{%
}{.}\hspace{.4pt}1778826}}


\bibitem{bennett2007timelapse}
\href{https://doi.org/https://doi.org/10.1145/1276377.1276505}{E.~P. Bennett
  and L.~McMillan}.
\newblock
  \href{https://doi.org/https://doi.org/10.1145/1276377.1276505}{Computational
  time-lapse video}.
\newblock \href{https://doi.org/https://doi.org/10.1145/1276377.1276505}{{\em
  ACM Trans. Gr.}},
  \href{https://doi.org/https://doi.org/10.1145/1276377.1276505}{26(3)},
  \href{https://doi.org/https://doi.org/10.1145/1276377.1276505}{2007}.
  \href{https://doi.org/10.1145/1276377.1276505}
{doi: \textsf{%
10\hspace{.1pt}\discretionary{.}{%
}{.}\hspace{.4pt}1145\discretionary{/}{%
}{/}1276377\hspace{.1pt}\discretionary{.}{%
}{.}\hspace{.4pt}1276505}}


\bibitem{borgo2011vidviz}
\href{https://doi.org/10.2312/EG2011/stars/001-023}{R.~Borgo, M.~Chen,
  B.~Daubney, E.~Grundy, G.~Heidemann, B.~Höferlin, M.~Höferlin, H.~Jänicke,
  D.~Weiskopf, and X.~Xie}.
\newblock \href{https://doi.org/10.2312/EG2011/stars/001-023}{{A Survey on
  Video-based Graphics and Video Visualization}}.
\newblock \href{https://doi.org/10.2312/EG2011/stars/001-023}{In N.~John and
  B.~Wyvill, eds., {\em Eurographics 2011 - State of the Art Reports}}.
  \href{https://doi.org/10.2312/EG2011/stars/001-023}{The Eurographics
  Association}, \href{https://doi.org/10.2312/EG2011/stars/001-023}{2011}.
  \href{https://doi.org/10.2312/EG2011/stars/001-023}
{doi: \textsf{%
10\hspace{.1pt}\discretionary{.}{%
}{.}\hspace{.4pt}2312\discretionary{/}{%
}{/}EG2011\discretionary{/}{%
}{/}stars\discretionary{/}{%
}{/}001\discretionary{%
}{-}{-}023}}


\bibitem{buckley2017ecology}
\href{https://doi.org/https://doi.org/10.5751/ES-09268-220330}{E.~M.~B.
  Buckley, C.~R. Allen, M.~L. Forsberg, M.~C. Farrell, and A.~J. Caven}.
\newblock
  \href{https://doi.org/https://doi.org/10.5751/ES-09268-220330}{Capturing
  change: the duality of time-lapse imagery to acquire data and depict
  ecological dynamics}.
\newblock \href{https://doi.org/https://doi.org/10.5751/ES-09268-220330}{{\em
  Ecology and Society}},
  \href{https://doi.org/https://doi.org/10.5751/ES-09268-220330}{22:1--12},
  \href{https://doi.org/https://doi.org/10.5751/ES-09268-220330}{2017}.
  \href{https://doi.org/10.5751/ES-09268-220330}
{doi: \textsf{%
10\hspace{.1pt}\discretionary{.}{%
}{.}\hspace{.4pt}5751\discretionary{/}{%
}{/}ES\discretionary{%
}{-}{-}09268\discretionary{%
}{-}{-}220330}}


\bibitem{burt1983laplacian}
\href{https://doi.org/10.1109/TCOM.1983.1095851}{P.~{Burt} and E.~{Adelson}}.
\newblock \href{https://doi.org/10.1109/TCOM.1983.1095851}{The laplacian
  pyramid as a compact image code}.
\newblock \href{https://doi.org/10.1109/TCOM.1983.1095851}{{\em IEEE
  Transactions on Communications}},
  \href{https://doi.org/10.1109/TCOM.1983.1095851}{31(4):532--540},
  \href{https://doi.org/10.1109/TCOM.1983.1095851}{1983}.
  \href{https://doi.org/10.1109/TCOM.1983.1095851}
{doi: \textsf{%
10\hspace{.1pt}\discretionary{.}{%
}{.}\hspace{.4pt}1109\discretionary{/}{%
}{/}TCOM\hspace{.1pt}\discretionary{.}{%
}{.}\hspace{.4pt}1983\hspace{.1pt}\discretionary{.}{%
}{.}\hspace{.4pt}1095851}}


\bibitem{burt1981pyramids}
\href{https://doi.org/https://doi.org/10.1016/0146-664X(81)90092-7}{P.~J.
  Burt}.
\newblock
  \href{https://doi.org/https://doi.org/10.1016/0146-664X(81)90092-7}{Fast
  filter transform for image processing}.
\newblock
  \href{https://doi.org/https://doi.org/10.1016/0146-664X(81)90092-7}{{\em
  Computer Graphics and Image Processing}},
  \href{https://doi.org/https://doi.org/10.1016/0146-664X(81)90092-7}{16(1):20--51},
  \href{https://doi.org/https://doi.org/10.1016/0146-664X(81)90092-7}{1981}.
  \href{https://doi.org/10.1016/0146-664X(81)90092-7}
{doi: \textsf{%
10\hspace{.1pt}\discretionary{.}{%
}{.}\hspace{.4pt}1016\discretionary{/}{%
}{/}0146\discretionary{%
}{-}{-}664X\discretionary{%
}{(}{(}81\discretionary{)}{%
}{)}90092\discretionary{%
}{-}{-}7}}


\bibitem{cakmak2021multiscale}
E.~{Cakmak}, U.~{Schlegel}, D.~{Jäckle}, D.~{Keim}, and T.~{Schreck}.
\newblock Multiscale snapshots: Visual analysis of temporal summaries in
  dynamic graphs.
\newblock {\em IEEE Transactions on Visualization and Computer Graphics},
  27(2):517--527, 2021.

\bibitem{cohen1995time}
L.~Cohen.
\newblock {\em Time-frequency analysis}, vol. 778.
\newblock Prentice-Hall, Inc., USA, 1995.

\bibitem{finkelstein1996multiresolution}
\href{https://doi.org/https://doi.org/10.1145/237170.237266}{A.~Finkelstein,
  C.~E. Jacobs, and D.~H. Salesin}.
\newblock
  \href{https://doi.org/https://doi.org/10.1145/237170.237266}{Multiresolution
  video}.
\newblock \href{https://doi.org/https://doi.org/10.1145/237170.237266}{{\em ACM
  Trans. Gr.}},
  \href{https://doi.org/https://doi.org/10.1145/237170.237266}{1996}.
  \href{https://doi.org/10.1145/237170.237266}
{doi: \textsf{%
10\hspace{.1pt}\discretionary{.}{%
}{.}\hspace{.4pt}1145\discretionary{/}{%
}{/}237170\hspace{.1pt}\discretionary{.}{%
}{.}\hspace{.4pt}237266}}


\bibitem{google2022timelapse}
Google.
\newblock {Exploring Timelapse in Google Earth}.
\newblock \url{https://www.youtube.com/watch?v=5W-zPqrGQWA}.

\bibitem{gutwin2019spreadloading}
\href{https://doi.org/https://doi.org/10.1145/3290605.3300785}{C.~Gutwin,
  M.~van~der Kamp, M.~S. Uddin, K.~Stanley, I.~Stavness, and S.~Vail}.
\newblock
  \href{https://doi.org/https://doi.org/10.1145/3290605.3300785}{Improving
  early navigation in time-lapse video with spread-frame loading}.
\newblock \href{https://doi.org/https://doi.org/10.1145/3290605.3300785}{In
  {\em Proceedings of the 2019 CHI Conference on Human Factors in Computing
  Systems}},
  \href{https://doi.org/https://doi.org/10.1145/3290605.3300785}{2019}.
  \href{https://doi.org/10.1145/3290605.3300785}
{doi: \textsf{%
10\hspace{.1pt}\discretionary{.}{%
}{.}\hspace{.4pt}1145\discretionary{/}{%
}{/}3290605\hspace{.1pt}\discretionary{.}{%
}{.}\hspace{.4pt}3300785}}


\bibitem{hansen2018art}
\href{https://doi.org/10.1080/14780887.2018.1430011}{S.~Hansen}.
\newblock \href{https://doi.org/10.1080/14780887.2018.1430011}{Video
  installation as a creative means of representing temporality in visual data}.
\newblock \href{https://doi.org/10.1080/14780887.2018.1430011}{{\em Qualitative
  Research in Psychology}},
  \href{https://doi.org/10.1080/14780887.2018.1430011}{15(2-3):292--297},
  \href{https://doi.org/10.1080/14780887.2018.1430011}{2018}.
  \href{https://doi.org/10.1080/14780887.2018.1430011}
{doi: \textsf{%
10\hspace{.1pt}\discretionary{.}{%
}{.}\hspace{.4pt}1080\discretionary{/}{%
}{/}14780887\hspace{.1pt}\discretionary{.}{%
}{.}\hspace{.4pt}2018\hspace{.1pt}\discretionary{.}{%
}{.}\hspace{.4pt}1430011}}


\bibitem{harrower2001animation}
\href{https://doi.org/10.14714/CP39.637}{M.~Harrower}.
\newblock \href{https://doi.org/10.14714/CP39.637}{Visualizing change: Using
  cartographic animation to explore remotely-sensed data}.
\newblock \href{https://doi.org/10.14714/CP39.637}{{\em Cartographic
  Perspectives}}, \href{https://doi.org/10.14714/CP39.637}{(39):30–42},
  \href{https://doi.org/10.14714/CP39.637}{Jun. 2001}.
  \href{https://doi.org/10.14714/CP39.637}
{doi: \textsf{%
10\hspace{.1pt}\discretionary{.}{%
}{.}\hspace{.4pt}14714\discretionary{/}{%
}{/}CP39\hspace{.1pt}\discretionary{.}{%
}{.}\hspace{.4pt}637}}


\bibitem{hartill2020fishing}
\href{https://doi.org/https://doi.org/10.1111/faf.12413}{B.~W. Hartill, S.~M.
  Taylor, K.~Keller, and M.~S. Weltersbach}.
\newblock \href{https://doi.org/https://doi.org/10.1111/faf.12413}{Digital
  camera monitoring of recreational fishing effort: Applications and
  challenges}.
\newblock \href{https://doi.org/https://doi.org/10.1111/faf.12413}{{\em Fish
  and Fisheries}},
  \href{https://doi.org/https://doi.org/10.1111/faf.12413}{21(1):204--215},
  \href{https://doi.org/https://doi.org/10.1111/faf.12413}{2020}.
  \href{https://doi.org/10.1111/faf.12413}
{doi: \textsf{%
10\hspace{.1pt}\discretionary{.}{%
}{.}\hspace{.4pt}1111\discretionary{/}{%
}{/}faf\hspace{.1pt}\discretionary{.}{%
}{.}\hspace{.4pt}12413}}


\bibitem{hoferlin2012ffvis}
\href{https://doi.org/10.1109/TVCG.2012.222}{M.~Höferlin, K.~Kurzhals,
  B.~Höferlin, G.~Heidemann, and D.~Weiskopf}.
\newblock \href{https://doi.org/10.1109/TVCG.2012.222}{Evaluation of
  fast-forward video visualization}.
\newblock \href{https://doi.org/10.1109/TVCG.2012.222}{{\em IEEE Transactions
  on Visualization and Computer Graphics}},
  \href{https://doi.org/10.1109/TVCG.2012.222}{18(12):2095--2103},
  \href{https://doi.org/10.1109/TVCG.2012.222}{2012}.
  \href{https://doi.org/10.1109/TVCG.2012.222}
{doi: \textsf{%
10\hspace{.1pt}\discretionary{.}{%
}{.}\hspace{.4pt}1109\discretionary{/}{%
}{/}TVCG\hspace{.1pt}\discretionary{.}{%
}{.}\hspace{.4pt}2012\hspace{.1pt}\discretionary{.}{%
}{.}\hspace{.4pt}222}}


\bibitem{jackson2013panopticon}
\href{https://doi.org/https://doi.org/10.1145/2501988.2502038}{D.~Jackson,
  J.~Nicholson, G.~Stoeckigt, R.~Wrobel, A.~Thieme, and P.~Olivier}.
\newblock
  \href{https://doi.org/https://doi.org/10.1145/2501988.2502038}{Panopticon: A
  parallel video overview system}.
\newblock \href{https://doi.org/https://doi.org/10.1145/2501988.2502038}{In
  {\em Proceedings of the 26th Annual ACM Symposium on User Interface Software
  and Technology}},
  \href{https://doi.org/https://doi.org/10.1145/2501988.2502038}{UIST '13}.
  \href{https://doi.org/https://doi.org/10.1145/2501988.2502038}{Association
  for Computing Machinery},
  \href{https://doi.org/https://doi.org/10.1145/2501988.2502038}{2013}.
  \href{https://doi.org/10.1145/2501988.2502038}
{doi: \textsf{%
10\hspace{.1pt}\discretionary{.}{%
}{.}\hspace{.4pt}1145\discretionary{/}{%
}{/}2501988\hspace{.1pt}\discretionary{.}{%
}{.}\hspace{.4pt}2502038}}


\bibitem{joshi2015hyperlapse}
\href{https://doi.org/https://doi.org/10.1145/2766954}{N.~Joshi, W.~Kienzle,
  M.~Toelle, M.~Uyttendaele, and M.~F. Cohen}.
\newblock \href{https://doi.org/https://doi.org/10.1145/2766954}{Real-time
  hyperlapse creation via optimal frame selection}.
\newblock \href{https://doi.org/https://doi.org/10.1145/2766954}{{\em ACM
  Trans. Gr.}}, \href{https://doi.org/https://doi.org/10.1145/2766954}{2015}.
  \href{https://doi.org/10.1145/2766954}
{doi: \textsf{%
10\hspace{.1pt}\discretionary{.}{%
}{.}\hspace{.4pt}1145\discretionary{/}{%
}{/}2766954}}


\bibitem{sierraclub2014}
M.~Kotack.
\newblock 4 environmental time-lapse videos you have to see to believe.
\newblock
  \url{https://www.sierraclub.org/sierra/2014-5-september-october/green-life/4-environmental-time-lapse-videos-you-have-see-believe},
  Oct 2014.

\bibitem{kratochvil2020somhunter}
\href{https://doi.org/10.1007/978-3-030-37734-2_71}{M.~Kratochv{\'i}l,
  P.~Vesel{\'y}, F.~Mejzl{\'i}k, and J.~Loko{\v{c}}}.
\newblock \href{https://doi.org/10.1007/978-3-030-37734-2_71}{{SOM}-hunter:
  Video browsing with relevance-to-{SOM} feedback loop}.
\newblock \href{https://doi.org/10.1007/978-3-030-37734-2_71}{In {\em
  MultiMedia Modeling}}.
  \href{https://doi.org/10.1007/978-3-030-37734-2_71}{Springer International
  Publishing}, \href{https://doi.org/10.1007/978-3-030-37734-2_71}{2020}.
  \href{https://doi.org/10.1007/978-3-030-37734-2_71}
{doi: \textsf{%
10\hspace{.1pt}\discretionary{.}{%
}{.}\hspace{.4pt}1007\discretionary{/}{%
}{/}978\discretionary{%
}{-}{-}3\discretionary{%
}{-}{-}030\discretionary{%
}{-}{-}37734\discretionary{%
}{-}{-}2\_71}}


\bibitem{lan2014tsp}
\href{https://doi.org/10.48550/ARXIV.1408.7071}{Z.~Lan, X.~Li, and A.~G.
  Hauptmann}.
\newblock \href{https://doi.org/10.48550/ARXIV.1408.7071}{Temporal extension of
  scale pyramid and spatial pyramid matching for action recognition},
  \href{https://doi.org/10.48550/ARXIV.1408.7071}{2014}.
  \href{https://doi.org/10.48550/ARXIV.1408.7071}
{doi: \textsf{%
10\hspace{.1pt}\discretionary{.}{%
}{.}\hspace{.4pt}48550\discretionary{/}{%
}{/}ARXIV\hspace{.1pt}\discretionary{.}{%
}{.}\hspace{.4pt}1408\hspace{.1pt}\discretionary{.}{%
}{.}\hspace{.4pt}7071}}


\bibitem{liu2016glacier}
\href{https://doi.org/https://doi.org/10.1007/s12665-015-5075-2}{J.~Liu,
  R.~Chen, C.~Han, and W.~Qing}.
\newblock
  \href{https://doi.org/https://doi.org/10.1007/s12665-015-5075-2}{Two-year
  comparative study of snow cover dynamics and its impact factors on glacier
  surface}.
\newblock \href{https://doi.org/https://doi.org/10.1007/s12665-015-5075-2}{{\em
  Environmental Earth Sciences}},
  \href{https://doi.org/https://doi.org/10.1007/s12665-015-5075-2}{75(3):1--11},
  \href{https://doi.org/https://doi.org/10.1007/s12665-015-5075-2}{2016}.
  \href{https://doi.org/10.1007/s12665-015-5075-2}
{doi: \textsf{%
10\hspace{.1pt}\discretionary{.}{%
}{.}\hspace{.4pt}1007\discretionary{/}{%
}{/}s12665\discretionary{%
}{-}{-}015\discretionary{%
}{-}{-}5075\discretionary{%
}{-}{-}2}}


\bibitem{lobo2019satellite}
\href{https://doi.org/10.1109/TVCG.2018.2796557}{M.-J. Lobo, C.~Appert, and
  E.~Pietriga}.
\newblock \href{https://doi.org/10.1109/TVCG.2018.2796557}{Animation plans for
  before-and-after satellite images}.
\newblock \href{https://doi.org/10.1109/TVCG.2018.2796557}{{\em IEEE
  Transactions on Visualization and Computer Graphics}},
  \href{https://doi.org/10.1109/TVCG.2018.2796557}{25(2):1347--1360},
  \href{https://doi.org/10.1109/TVCG.2018.2796557}{2019}.
  \href{https://doi.org/10.1109/TVCG.2018.2796557}
{doi: \textsf{%
10\hspace{.1pt}\discretionary{.}{%
}{.}\hspace{.4pt}1109\discretionary{/}{%
}{/}TVCG\hspace{.1pt}\discretionary{.}{%
}{.}\hspace{.4pt}2018\hspace{.1pt}\discretionary{.}{%
}{.}\hspace{.4pt}2796557}}


\bibitem{martinbrualla2015timelapse}
\href{https://doi.org/https://doi.org/10.1145/2766903}{R.~Martin-Brualla,
  D.~Gallup, and S.~M. Seitz}.
\newblock \href{https://doi.org/https://doi.org/10.1145/2766903}{Time-lapse
  mining from internet photos}.
\newblock \href{https://doi.org/https://doi.org/10.1145/2766903}{{\em ACM
  Trans. Gr.}}, \href{https://doi.org/https://doi.org/10.1145/2766903}{2015}.
  \href{https://doi.org/10.1145/2766903}
{doi: \textsf{%
10\hspace{.1pt}\discretionary{.}{%
}{.}\hspace{.4pt}1145\discretionary{/}{%
}{/}2766903}}


\bibitem{monea2021education}
\href{https://doi.org/10.1177/14687941211019524}{B.~Monea, A.~Stornaiuolo, and
  E.~P. Catena}.
\newblock \href{https://doi.org/10.1177/14687941211019524}{Reframing
  temporality in participatory visual research with timelapse video}.
\newblock \href{https://doi.org/10.1177/14687941211019524}{{\em Qualitative
  Research}}, \href{https://doi.org/10.1177/14687941211019524}{2021}.
  \href{https://doi.org/10.1177/14687941211019524}
{doi: \textsf{%
10\hspace{.1pt}\discretionary{.}{%
}{.}\hspace{.4pt}1177\discretionary{/}{%
}{/}14687941211019524}}


\bibitem{nakamura2019education}
\href{https://doi.org/10.3390/educsci9030190}{K.~W. Nakamura, A.~Fujiwara,
  H.~H. Kobayashi, and K.~Saito}.
\newblock \href{https://doi.org/10.3390/educsci9030190}{Multi-timescale
  education program for temporal expansion in ecocentric education: Using
  fixed-point time-lapse images for phenology observation}.
\newblock \href{https://doi.org/10.3390/educsci9030190}{{\em Education
  Sciences}}, \href{https://doi.org/10.3390/educsci9030190}{9(3)},
  \href{https://doi.org/10.3390/educsci9030190}{2019}.
  \href{https://doi.org/10.3390/educsci9030190}
{doi: \textsf{%
10\hspace{.1pt}\discretionary{.}{%
}{.}\hspace{.4pt}3390\discretionary{/}{%
}{/}educsci9030190}}


\bibitem{nayak2021anomaly}
\href{https://doi.org/https://doi.org/10.1016/j.imavis.2020.104078}{R.~Nayak,
  U.~C. Pati, and S.~K. Das}.
\newblock \href{https://doi.org/https://doi.org/10.1016/j.imavis.2020.104078}{A
  comprehensive review on deep learning-based methods for video anomaly
  detection}.
\newblock
  \href{https://doi.org/https://doi.org/10.1016/j.imavis.2020.104078}{{\em
  Image and Vision Computing}},
  \href{https://doi.org/https://doi.org/10.1016/j.imavis.2020.104078}{106:104078},
  \href{https://doi.org/https://doi.org/10.1016/j.imavis.2020.104078}{2021}.
  \href{https://doi.org/10.1016/j.imavis.2020.104078}
{doi: \textsf{%
10\hspace{.1pt}\discretionary{.}{%
}{.}\hspace{.4pt}1016\discretionary{/}{%
}{/}j\hspace{.1pt}\discretionary{.}{%
}{.}\hspace{.4pt}imavis\hspace{.1pt}\discretionary{.}{%
}{.}\hspace{.4pt}2020\hspace{.1pt}\discretionary{.}{%
}{.}\hspace{.4pt}104078}}


\bibitem{pritch2008nonchronological}
\href{https://doi.org/10.1109/TPAMI.2008.29}{Y.~Pritch, A.~Rav-Acha, and
  S.~Peleg}.
\newblock \href{https://doi.org/10.1109/TPAMI.2008.29}{Nonchronological video
  synopsis and indexing}.
\newblock \href{https://doi.org/10.1109/TPAMI.2008.29}{{\em IEEE Transactions
  on Pattern Analysis and Machine Intelligence}},
  \href{https://doi.org/10.1109/TPAMI.2008.29}{30(11):1971--1984},
  \href{https://doi.org/10.1109/TPAMI.2008.29}{2008}.
  \href{https://doi.org/10.1109/TPAMI.2008.29}
{doi: \textsf{%
10\hspace{.1pt}\discretionary{.}{%
}{.}\hspace{.4pt}1109\discretionary{/}{%
}{/}TPAMI\hspace{.1pt}\discretionary{.}{%
}{.}\hspace{.4pt}2008\hspace{.1pt}\discretionary{.}{%
}{.}\hspace{.4pt}29}}


\bibitem{rav-acha05evolving}
\href{https://www.vision.huji.ac.il/videowarping/}{A.~Rav-acha, Y.~Pritch,
  D.~Lischinski, and S.~Peleg}.
\newblock \href{https://www.vision.huji.ac.il/videowarping/}{Evolving time
  fronts: Spatio-temporal video warping}.
\newblock \href{https://www.vision.huji.ac.il/videowarping/}{Technical report},
  \href{https://www.vision.huji.ac.il/videowarping/}{2005}.

\bibitem{rochan2018video}
\href{https://doi.org/10.1007/978-3-030-01258-8_22}{M.~Rochan, L.~Ye, and
  Y.~Wang}.
\newblock \href{https://doi.org/10.1007/978-3-030-01258-8_22}{Video
  summarization using fully convolutional sequence networks}.
\newblock \href{https://doi.org/10.1007/978-3-030-01258-8_22}{In {\em
  Proceedings of the European Conference on Computer Vision (ECCV)}},
  \href{https://doi.org/10.1007/978-3-030-01258-8_22}{pp. 347--363},
  \href{https://doi.org/10.1007/978-3-030-01258-8_22}{2018}.
  \href{https://doi.org/10.1007/978-3-030-01258-8_22}
{doi: \textsf{%
10\hspace{.1pt}\discretionary{.}{%
}{.}\hspace{.4pt}1007\discretionary{/}{%
}{/}978\discretionary{%
}{-}{-}3\discretionary{%
}{-}{-}030\discretionary{%
}{-}{-}01258\discretionary{%
}{-}{-}8\_22}}


\bibitem{romero2008vizavis}
\href{https://doi.org/10.1109/TVCG.2008.185}{M.~{Romero}, J.~{Summet},
  J.~{Stasko}, and G.~{Abowd}}.
\newblock \href{https://doi.org/10.1109/TVCG.2008.185}{Viz-a-vis: Toward
  visualizing video through computer vision}.
\newblock \href{https://doi.org/10.1109/TVCG.2008.185}{{\em IEEE Transactions
  on Visualization and Computer Graphics}},
  \href{https://doi.org/10.1109/TVCG.2008.185}{14(6):1261--1268},
  \href{https://doi.org/10.1109/TVCG.2008.185}{2008}.
  \href{https://doi.org/10.1109/TVCG.2008.185}
{doi: \textsf{%
10\hspace{.1pt}\discretionary{.}{%
}{.}\hspace{.4pt}1109\discretionary{/}{%
}{/}TVCG\hspace{.1pt}\discretionary{.}{%
}{.}\hspace{.4pt}2008\hspace{.1pt}\discretionary{.}{%
}{.}\hspace{.4pt}185}}


\bibitem{rossetto2020vbs}
\href{https://doi.org/10.1109/TMM.2020.2980944}{L.~{Rossetto}, R.~{Gasser},
  J.~{Lokoč}, W.~{Bailer}, K.~{Schoeffmann}, B.~{Muenzer}, T.~{Souček}, P.~A.
  {Nguyen}, P.~{Bolettieri}, A.~{Leibetseder}, and S.~{Vrochidis}}.
\newblock \href{https://doi.org/10.1109/TMM.2020.2980944}{Interactive video
  retrieval in the age of deep learning – detailed evaluation of {VBS} 2019}.
\newblock \href{https://doi.org/10.1109/TMM.2020.2980944}{{\em IEEE
  Transactions on Multimedia}},
  \href{https://doi.org/10.1109/TMM.2020.2980944}{23:243--256},
  \href{https://doi.org/10.1109/TMM.2020.2980944}{2021}.
  \href{https://doi.org/10.1109/TMM.2020.2980944}
{doi: \textsf{%
10\hspace{.1pt}\discretionary{.}{%
}{.}\hspace{.4pt}1109\discretionary{/}{%
}{/}TMM\hspace{.1pt}\discretionary{.}{%
}{.}\hspace{.4pt}2020\hspace{.1pt}\discretionary{.}{%
}{.}\hspace{.4pt}2980944}}


\bibitem{rubinstein2011motion}
\href{https://doi.org/10.1109/CVPR.2011.5995374}{M.~Rubinstein, C.~Liu,
  P.~Sand, F.~Durand, and W.~T. Freeman}.
\newblock \href{https://doi.org/10.1109/CVPR.2011.5995374}{Motion denoising
  with application to time-lapse photography}.
\newblock \href{https://doi.org/10.1109/CVPR.2011.5995374}{In {\em Proceedings
  of the 2011 IEEE Conference on Computer Vision and Pattern Recognition}},
  \href{https://doi.org/10.1109/CVPR.2011.5995374}{CVPR '11},
  \href{https://doi.org/10.1109/CVPR.2011.5995374}{p. 313–320}.
  \href{https://doi.org/10.1109/CVPR.2011.5995374}{IEEE Computer Society},
  \href{https://doi.org/10.1109/CVPR.2011.5995374}{USA},
  \href{https://doi.org/10.1109/CVPR.2011.5995374}{2011}.
  \href{https://doi.org/10.1109/CVPR.2011.5995374}
{doi: \textsf{%
10\hspace{.1pt}\discretionary{.}{%
}{.}\hspace{.4pt}1109\discretionary{/}{%
}{/}CVPR\hspace{.1pt}\discretionary{.}{%
}{.}\hspace{.4pt}2011\hspace{.1pt}\discretionary{.}{%
}{.}\hspace{.4pt}5995374}}


\bibitem{seyednasrollah2019phenology}
\href{https://doi.org/10.1038/s41597-019-0229-9}{B.~Seyednasrollah, A.~M.
  Young, K.~Hufkens, T.~Milliman, M.~A. Friedl, S.~Frolking, and A.~D.
  Richardson}.
\newblock \href{https://doi.org/10.1038/s41597-019-0229-9}{Tracking vegetation
  phenology across diverse biomes using version 2.0 of the phenocam dataset.}
\newblock \href{https://doi.org/10.1038/s41597-019-0229-9}{{\em Scientific
  Data}}, \href{https://doi.org/10.1038/s41597-019-0229-9}{6(1)},
  \href{https://doi.org/10.1038/s41597-019-0229-9}{2019}.
  \href{https://doi.org/10.1038/s41597-019-0229-9}
{doi: \textsf{%
10\hspace{.1pt}\discretionary{.}{%
}{.}\hspace{.4pt}1038\discretionary{/}{%
}{/}s41597\discretionary{%
}{-}{-}019\discretionary{%
}{-}{-}0229\discretionary{%
}{-}{-}9}}


\bibitem{shao2014actionrecog}
\href{https://doi.org/10.1109/TCYB.2013.2273174}{L.~Shao, X.~Zhen, D.~Tao, and
  X.~Li}.
\newblock \href{https://doi.org/10.1109/TCYB.2013.2273174}{Spatio-temporal
  laplacian pyramid coding for action recognition}.
\newblock \href{https://doi.org/10.1109/TCYB.2013.2273174}{{\em IEEE
  Transactions on Cybernetics}},
  \href{https://doi.org/10.1109/TCYB.2013.2273174}{44(6):817--827},
  \href{https://doi.org/10.1109/TCYB.2013.2273174}{2014}.
  \href{https://doi.org/10.1109/TCYB.2013.2273174}
{doi: \textsf{%
10\hspace{.1pt}\discretionary{.}{%
}{.}\hspace{.4pt}1109\discretionary{/}{%
}{/}TCYB\hspace{.1pt}\discretionary{.}{%
}{.}\hspace{.4pt}2013\hspace{.1pt}\discretionary{.}{%
}{.}\hspace{.4pt}2273174}}


\bibitem{schneiderman1996mantra}
\href{https://doi.org/10.1109/VL.1996.545307}{B.~Shneiderman}.
\newblock \href{https://doi.org/10.1109/VL.1996.545307}{The eyes have it: a
  task by data type taxonomy for information visualizations}.
\newblock \href{https://doi.org/10.1109/VL.1996.545307}{In {\em Proceedings
  1996 IEEE Symposium on Visual Languages}},
  \href{https://doi.org/10.1109/VL.1996.545307}{pp. 336--343},
  \href{https://doi.org/10.1109/VL.1996.545307}{1996}.
  \href{https://doi.org/10.1109/VL.1996.545307}
{doi: \textsf{%
10\hspace{.1pt}\discretionary{.}{%
}{.}\hspace{.4pt}1109\discretionary{/}{%
}{/}VL\hspace{.1pt}\discretionary{.}{%
}{.}\hspace{.4pt}1996\hspace{.1pt}\discretionary{.}{%
}{.}\hspace{.4pt}545307}}


\bibitem{szeliski2007image}
\href{https://doi.org/10.1561/0600000009}{R.~Szeliski et~al.}
\newblock \href{https://doi.org/10.1561/0600000009}{Image alignment and
  stitching: A tutorial}.
\newblock \href{https://doi.org/10.1561/0600000009}{{\em Foundations and
  Trends{\textregistered} in Computer Graphics and Vision}},
  \href{https://doi.org/10.1561/0600000009}{2(1):1--104},
  \href{https://doi.org/10.1561/0600000009}{2007}.
  \href{https://doi.org/10.1561/0600000009}
{doi: \textsf{%
10\hspace{.1pt}\discretionary{.}{%
}{.}\hspace{.4pt}1561\discretionary{/}{%
}{/}0600000009}}


\bibitem{temponaut2021vid}
Temponaut.
\newblock {TEMPONAUT Top 10 Timelapses 2021}.
\newblock \url{https://www.youtube.com/watch?v=Sw60Ehm1b5c}.

\bibitem{tibaut2018construction}
\href{https://doi.org/https://doi.org/10.1007/s11625-018-0595-9}{A.~Tibaut and
  D.~Zazula}.
\newblock
  \href{https://doi.org/https://doi.org/10.1007/s11625-018-0595-9}{Sustainable
  management of construction site big visual data}.
\newblock \href{https://doi.org/https://doi.org/10.1007/s11625-018-0595-9}{{\em
  Sustainability Science}},
  \href{https://doi.org/https://doi.org/10.1007/s11625-018-0595-9}{13(5):1311--1322},
  \href{https://doi.org/https://doi.org/10.1007/s11625-018-0595-9}{2018}.
  \href{https://doi.org/10.1007/s11625-018-0595-9}
{doi: \textsf{%
10\hspace{.1pt}\discretionary{.}{%
}{.}\hspace{.4pt}1007\discretionary{/}{%
}{/}s11625\discretionary{%
}{-}{-}018\discretionary{%
}{-}{-}0595\discretionary{%
}{-}{-}9}}


\bibitem{usgs2016repeat}
{U}nited {S}tates~{G}eological {S}urvey.
\newblock {Repeat Photography Project}.
\newblock
  \url{https://www.usgs.gov/centers/norock/science/repeat-photography-project},
  {2016}.

\bibitem{vollmer2018education}
\href{https://doi.org/10.1088/1361-6552/aaa954}{M.~Vollmer and K.-P.
  Möllmann}.
\newblock \href{https://doi.org/10.1088/1361-6552/aaa954}{Slow
  speed{\textemdash}fast motion: time-lapse recordings in physics education}.
\newblock \href{https://doi.org/10.1088/1361-6552/aaa954}{{\em Physics
  Education}}, \href{https://doi.org/10.1088/1361-6552/aaa954}{53(3):035019},
  \href{https://doi.org/10.1088/1361-6552/aaa954}{March 2018}.
  \href{https://doi.org/10.1088/1361-6552/aaa954}
{doi: \textsf{%
10\hspace{.1pt}\discretionary{.}{%
}{.}\hspace{.4pt}1088\discretionary{/}{%
}{/}1361\discretionary{%
}{-}{-}6552\discretionary{/}{%
}{/}aaa954}}


\bibitem{wang2017actionpooling}
\href{https://doi.org/10.1109/TCSVT.2016.2576761}{P.~Wang, Y.~Cao, C.~Shen,
  L.~Liu, and H.~T. Shen}.
\newblock \href{https://doi.org/10.1109/TCSVT.2016.2576761}{Temporal pyramid
  pooling-based convolutional neural network for action recognition}.
\newblock \href{https://doi.org/10.1109/TCSVT.2016.2576761}{{\em IEEE
  Transactions on Circuits and Systems for Video Technology}},
  \href{https://doi.org/10.1109/TCSVT.2016.2576761}{27(12):2613--2622},
  \href{https://doi.org/10.1109/TCSVT.2016.2576761}{2017}.
  \href{https://doi.org/10.1109/TCSVT.2016.2576761}
{doi: \textsf{%
10\hspace{.1pt}\discretionary{.}{%
}{.}\hspace{.4pt}1109\discretionary{/}{%
}{/}TCSVT\hspace{.1pt}\discretionary{.}{%
}{.}\hspace{.4pt}2016\hspace{.1pt}\discretionary{.}{%
}{.}\hspace{.4pt}2576761}}


\bibitem{wang2017action}
\href{https://doi.org/10.1109/CVPR.2017.226}{Y.~Wang, M.~Long, J.~Wang, and
  P.~S. Yu}.
\newblock \href{https://doi.org/10.1109/CVPR.2017.226}{Spatiotemporal pyramid
  network for video action recognition}.
\newblock \href{https://doi.org/10.1109/CVPR.2017.226}{In {\em 2017 IEEE
  Conference on Computer Vision and Pattern Recognition (CVPR)}},
  \href{https://doi.org/10.1109/CVPR.2017.226}{pp. 2097--2106},
  \href{https://doi.org/10.1109/CVPR.2017.226}{2017}.
  \href{https://doi.org/10.1109/CVPR.2017.226}
{doi: \textsf{%
10\hspace{.1pt}\discretionary{.}{%
}{.}\hspace{.4pt}1109\discretionary{/}{%
}{/}CVPR\hspace{.1pt}\discretionary{.}{%
}{.}\hspace{.4pt}2017\hspace{.1pt}\discretionary{.}{%
}{.}\hspace{.4pt}226}}


\bibitem{wehrwein2021scene}
\href{https://doi.org/10.1109/TVCG.2020.2993195}{S.~{Wehrwein}, K.~{Bala}, and
  N.~{Snavely}}.
\newblock \href{https://doi.org/10.1109/TVCG.2020.2993195}{Scene summarization
  via motion normalization}.
\newblock \href{https://doi.org/10.1109/TVCG.2020.2993195}{{\em IEEE
  Transactions on Visualization and Computer Graphics}},
  \href{https://doi.org/10.1109/TVCG.2020.2993195}{27(4):2495--2501},
  \href{https://doi.org/10.1109/TVCG.2020.2993195}{2021}.
  \href{https://doi.org/10.1109/TVCG.2020.2993195}
{doi: \textsf{%
10\hspace{.1pt}\discretionary{.}{%
}{.}\hspace{.4pt}1109\discretionary{/}{%
}{/}TVCG\hspace{.1pt}\discretionary{.}{%
}{.}\hspace{.4pt}2020\hspace{.1pt}\discretionary{.}{%
}{.}\hspace{.4pt}2993195}}


\bibitem{wikipedia2022timelapse}
{Wikipedia contributors}.
\newblock Time-lapse photography --- {Wikipedia}{,} the free encyclopedia.
\newblock
  \url{https://en.wikipedia.org/w/index.php?title=Time-lapse_photography},
  2022.

\bibitem{yang2015construction}
\href{https://doi.org/https://doi.org/10.1016/j.aei.2015.01.011}{J.~Yang, M.-W.
  Park, P.~A. Vela, and M.~Golparvar-Fard}.
\newblock
  \href{https://doi.org/https://doi.org/10.1016/j.aei.2015.01.011}{Construction
  performance monitoring via still images, time-lapse photos, and video
  streams: Now, tomorrow, and the future}.
\newblock \href{https://doi.org/https://doi.org/10.1016/j.aei.2015.01.011}{{\em
  Advanced Engineering Informatics}},
  \href{https://doi.org/https://doi.org/10.1016/j.aei.2015.01.011}{29(2):211--224},
  \href{https://doi.org/https://doi.org/10.1016/j.aei.2015.01.011}{2015}.
\newblock
  \href{https://doi.org/https://doi.org/10.1016/j.aei.2015.01.011}{Infrastructure
  Computer Vision}. \href{https://doi.org/10.1016/j.aei.2015.01.011}
{doi: \textsf{%
10\hspace{.1pt}\discretionary{.}{%
}{.}\hspace{.4pt}1016\discretionary{/}{%
}{/}j\hspace{.1pt}\discretionary{.}{%
}{.}\hspace{.4pt}aei\hspace{.1pt}\discretionary{.}{%
}{.}\hspace{.4pt}2015\hspace{.1pt}\discretionary{.}{%
}{.}\hspace{.4pt}01\hspace{.1pt}\discretionary{.}{%
}{.}\hspace{.4pt}011}}


\bibitem{zhang2016video}
\href{https://doi.org/https://doi.org/10.1007/978-3-319-46478-7_47}{K.~Zhang,
  W.-L. Chao, F.~Sha, and K.~Grauman}.
\newblock
  \href{https://doi.org/https://doi.org/10.1007/978-3-319-46478-7_47}{Video
  summarization with long short-term memory}.
\newblock
  \href{https://doi.org/https://doi.org/10.1007/978-3-319-46478-7_47}{In {\em
  ECCV}},
  \href{https://doi.org/https://doi.org/10.1007/978-3-319-46478-7_47}{2016}.
  \href{https://doi.org/10.1007/978-3-319-46478-7_47}
{doi: \textsf{%
10\hspace{.1pt}\discretionary{.}{%
}{.}\hspace{.4pt}1007\discretionary{/}{%
}{/}978\discretionary{%
}{-}{-}3\discretionary{%
}{-}{-}319\discretionary{%
}{-}{-}46478\discretionary{%
}{-}{-}7\_47}}


\bibitem{zhang2017photometric}
\href{https://doi.org/https://doi.org/10.1111/cgf.13276}{X.~Zhang, J.-Y. Lee,
  K.~Sunkavalli, and Z.~Wang}.
\newblock \href{https://doi.org/https://doi.org/10.1111/cgf.13276}{Photometric
  stabilization for fast-forward videos}.
\newblock \href{https://doi.org/https://doi.org/10.1111/cgf.13276}{{\em
  Computer Graphics Forum}},
  \href{https://doi.org/https://doi.org/10.1111/cgf.13276}{36(7):105--113},
  \href{https://doi.org/https://doi.org/10.1111/cgf.13276}{2017}.
  \href{https://doi.org/10.1111/cgf.13276}
{doi: \textsf{%
10\hspace{.1pt}\discretionary{.}{%
}{.}\hspace{.4pt}1111\discretionary{/}{%
}{/}cgf\hspace{.1pt}\discretionary{.}{%
}{.}\hspace{.4pt}13276}}


\bibitem{zheng2019action}
\href{https://doi.org/https://doi.org/10.1016/j.neucom.2019.05.058}{Z.~Zheng,
  G.~An, D.~Wu, and Q.~Ruan}.
\newblock
  \href{https://doi.org/https://doi.org/10.1016/j.neucom.2019.05.058}{Spatial-temporal
  pyramid based convolutional neural network for action recognition}.
\newblock
  \href{https://doi.org/https://doi.org/10.1016/j.neucom.2019.05.058}{{\em
  Neurocomputing}},
  \href{https://doi.org/https://doi.org/10.1016/j.neucom.2019.05.058}{358:446--455},
  \href{https://doi.org/https://doi.org/10.1016/j.neucom.2019.05.058}{2019}.
  \href{https://doi.org/10.1016/j.neucom.2019.05.058}
{doi: \textsf{%
10\hspace{.1pt}\discretionary{.}{%
}{.}\hspace{.4pt}1016\discretionary{/}{%
}{/}j\hspace{.1pt}\discretionary{.}{%
}{.}\hspace{.4pt}neucom\hspace{.1pt}\discretionary{.}{%
}{.}\hspace{.4pt}2019\hspace{.1pt}\discretionary{.}{%
}{.}\hspace{.4pt}05\hspace{.1pt}\discretionary{.}{%
}{.}\hspace{.4pt}058}}


\bibitem{zhou2014time}
\href{https://doi.org/10.1109/CVPR.2014.429}{F.~Zhou, S.~B. Kang, and
  M.~Cohen}.
\newblock \href{https://doi.org/10.1109/CVPR.2014.429}{Time-mapping using
  space-time saliency}.
\newblock \href{https://doi.org/10.1109/CVPR.2014.429}{In {\em CVPR}},
  \href{https://doi.org/10.1109/CVPR.2014.429}{2014}.
  \href{https://doi.org/10.1109/CVPR.2014.429}
{doi: \textsf{%
10\hspace{.1pt}\discretionary{.}{%
}{.}\hspace{.4pt}1109\discretionary{/}{%
}{/}CVPR\hspace{.1pt}\discretionary{.}{%
}{.}\hspace{.4pt}2014\hspace{.1pt}\discretionary{.}{%
}{.}\hspace{.4pt}429}}


\end{thebibliography}
